\documentclass[10pt, conference, compsocconf]{IEEEtran}
\usepackage[utf8]{inputenc} 
\usepackage[T1]{fontenc}    
\usepackage{url}            
\usepackage{booktabs}       
\usepackage{amsfonts}       
\usepackage{nicefrac}       
\usepackage{microtype}      

\usepackage{amssymb}
\usepackage{amsmath}
\usepackage{graphicx} 
\usepackage{times}    
\usepackage[lofdepth,lotdepth]{subfig}
\usepackage{multirow}
\usepackage{bbm}
\usepackage{pseudocode}
\usepackage{color}
\usepackage{cite}
\usepackage{multirow}
\usepackage{bbm}
\usepackage{pseudocode}
\usepackage{algpseudocode}
\usepackage[procnumbered,ruled]{algorithm2e}
\usepackage{comment}
\usepackage[misc]{ifsym}

\usepackage{xcolor}
\usepackage{tabu}
\usepackage{colortbl}
\SetKwInput{Parameter}{Parameter}
\SetKwInput{Input}{Input}
\SetKwInput{Output}{Output}

\definecolor{kugray5}{RGB}{224,224,224}

\usepackage{booktabs} 
\newcommand{\TheName}{\texttt{SecureGBM}}

\begin{document}
%
\title{\TheName: Secure Multi-Party Gradient Boosting} 
\author{Zhi Feng$^{\dag,*}$, Haoyi Xiong$^{\dag,*}$, Chuanyuan Song$^\dag$, Sijia Yang$^{\ddag}$, Baoxin Zhao$^{\dag,\#}$,\\ Licheng Wang$^{\ddag}$, Zeyu Chen$^\dag$, Shengwen Yang$^\dag$, Liping Liu$^\dag$ and Jun Huan$^\dag$\\
$^\dag$ Big Data Group (BDG), Big Data Lab (BDL) and PaddlePaddle (DLTP), Baidu Inc., Beijing, China\\
$^\ddag$ State Key Laboratory of Networking and Switching Technology,\\
 Beijing University of Posts and Telecommunications, Beijing, China\\
$^\#$ Shenzhen Institutes of Advanced Technology,
Chinese Academy of Sciences, Shenzhen, China
}

\maketitle

\begin{abstract}
Federated machine learning systems have been widely used to facilitate the joint data analytics across the distributed datasets owned by the different parties that do not trust each others. In this paper, we proposed a novel Gradient Boosting Machines (GBM) framework \TheName\ built-up with a multi-party computation model based on semi-homomorphic encryption, where every involved party can jointly obtain a shared Gradient Boosting machines model while protecting their own data from the potential privacy leakage and inferential identification. More specific, our work focused on a specific ``dual-party''secure learning scenario based on two parties --- \emph{both party own an unique view (i.e., attributes or features) to the sample group of samples while only one party owns the labels}. In such scenario, feature and label data are not allowed to share with others. 

To achieve the above goal, we firstly extent --- LightGBM --- a well known implementation of tree-based GBM through covering its key operations for training and inference with SEAL homomorphic encryption schemes. However, the performance of such re-implementation is significantly bottle-necked by the explosive inflation of the communication payloads, based on ciphertexts subject to the increasing length of plaintexts. In this way, we then proposed to use stochastic approximation techniques to reduced the communication payloads while accelerating the overall training procedure in a statistical manner.
Our experiments using the real-world data showed that 
%
\TheName\ can well secure the communication and computation of LightGBM training and inference procedures for the both parties while only losing less than 3\% AUC, using the same number of iterations for gradient boosting, on a wide range of benchmark datasets. More specific, compared to LightGBM, the proposed \TheName\ would slowdown with 3x $\sim$ 64x time consumption per iteration in the training procedure, while \TheName\ becomes more and more efficient when the scale of the training dataset increases (i.e., the larger training set, the lower slowdown ratio).~\footnote{~$^*$Equal Contribution.~The manuscript has been accepted for publication at IEEE BigData 2019.}
\end{abstract}

\section{Introduction}
Multi-Party federated learning~\cite{chen2006algebraic} becomes one of the most popular machine learning paradigm thanks to the increasing trends of distributed data collection, storage and processing, as well as its benefits to the privacy-preserved manner in different kinds of applications. In most multi-party machine learning applications,  ``no raw data sharing'' is an important pre-condition, where the model should be trained using all data stored in distributed machines (i.e., parties) without any cross-machine raw data sharing. A wide range of machine learning models and algorithms, including logistic regression~\cite{pathak2010multiparty}, sparse discriminant analysis~\cite{bian2017multi,tian2016communication}, stochastic gradient-based learners~\cite{jayaraman2018distributed,xing2015petuum,ormandi2013gossip}, have been re-implemented on distributed computing, encryption, and privacy preserving computation/communication platforms, so as to incorporate the secure computation paradigms~\cite{chen2006algebraic}.

\textbf{Backgrounds and Related Work.} Existing efforts majorly work on the implementation of efficient federated learning systems. Two parallel computation paradigms---\emph{data-centric} and \emph{model-centric}~\cite{xing2015petuum,zhou2008large,dean2012large,tsianos2012consensus,tsianos2012consensus,smyth2009asynchronous,ormandi2013gossip} have been proposed. On each machine, the data centric algorithm first estimates the same set of parameters (of the model) using the local data, then aggregates the estimated parameters via model-averaging for global estimation. The model with aggregated parameters is considered as the trained model based on the overall data (from multiple parties) and before aggregated these parameters can be estimated through parallel computing structure in an easy way. Meanwhile, model-centric algorithms require multiple machines to share the same loss function with ``updatable parameters'', and allow each machine to update the parameters in the loss function using the local data so as to minimize the loss. Based on this characteristic, model-centric algorithm commonly updates the parameters sequentially so that the additional time consumption in updating is sometimes a tough nut for specific applications. Even so, compared with the data-centric, the model-centric methods usually can achieve better performances, as it minimizes the risk of the model~\cite{xing2015petuum,ormandi2013gossip}. To advance the distributed performance of linear classifiers, Tian et al.~\cite{tian2016communication} proposed a data-centric sparse linear discriminant analysis algorithm, which leverages the advantage of parallel computing. 

In terms of multi-party collaboration, the federated learning algorithms can be categorized into two types: \emph{Data separation} and \emph{View separation}. For the data separation, the algorithms are assumed to learn from the distributed datasets, where each dataset consists of a subset of samples of the same types~\cite{xing2015petuum,bian2017multi,tian2016communication,jayaraman2018distributed}. For example, hospitals are usually required to collaboratively learn a model to predict patents' future diseases through classifying their electronic medical records, where all hospitals follows the same scheme to collect patients' medical record while every hospital can only cover a part of the patients. In this case, federated learning here improves the overall performance of learning through incorporating the private datasets owned by different parties, while ensuring the privacy and security~\cite{xing2015petuum,ormandi2013gossip,jayaraman2018distributed}. While the existing data/computation parallelism mechanisms were usually motivated to improve federated learning under the data separation settings, the federated learning systems under the \emph{view separation} settings are seldom considered.%

\textbf{Our Work.} We mainly focus on \emph{view separation} settings of the federated learning that assumes the data view of the same group of samples are separated by multiple parties who do not trust each other. For example, the healthcare, finance, and insurance records of the same group of healthcare users are usually stored in the data centers of healthcare providers, banks, and insurance companies separately. For the healthcare users, they usually need some recommendations on the healthcare insurance products according to their health and financial status, while healthcare insurance companies need to learn from large-scale healthcare together with personal financial data to build such recommendation models. However, according to
the law enforcement about data privacy, it is difficult for these three partities to share their data with each other and learn such a predictive model. In this way, federated learning under view separation models is highly appreciated. In this work, we aim at working on the view separation federated learning algorithms using Gradient Boosting Machines (GBM) as the Classifiers. GBM is studied here as it can deliver decent prediction results and be interpreted by human experts for joint data analytics and cross-institutes data understanding purposes.

\textbf{Our Contributions.} We summarize the contribution of the proposed \TheName{} algorithm in following aspects. 
\begin{itemize}
\item Firstly, we study and formulate the federated learning problem under the (semi)-homomorphic encryption settings, while assuming the data owned by two parties are not sharable. More specific, in this paper, we assume each party owns a unique private view to the same group of samples, while the labels of these samples are monopolized by one party. To the best of our knowledge, this is the first study on tree-based Gradient Boosting Machine classifiers, by addressing 1) two-party security constraint, 2) efficient \emph{model-centric} learning with views separated by two parties but labels ``monopolied'' by one, and 3) the trade-off between statistical accuracy and the communication cost caused by statistical learning over encrypted communication. 

\item Secondly, to achieve the goals, we design the \TheName{} algorithm which re-implements the vanilla gradient-boosting tree based learners using semi-homomorphic encrypted computation operators offered by Microsoft SEAL. More specific, \TheName\ first replaces the \emph{addition} and \emph{multiplication} operators used in the gradient-boosting trees with the secured operators based on semi-homomorphic computation, then \TheName\ re-designs a new set of \emph{binary comparison} operators (i.e., $\geq$ or $\leq$) which can not be intercepted by attackers to exactly recover the ground truth through searching with the comparison operators (e.g., binary search). 

\item Furthermore, we observe the trade-off between statistical accuracy and communication cost for GBM training. One can use stochastic gradient boosting mechanism to update the training model with mini-batch of data per round, while the communication cost per round can be significantly reduced in a quadratics manner. However, compared to vanilla gradient boosting machines, additional rounds of training procedure might be needed by such stochastic gradient boosting to achieve equivilent performance. In this way, \TheName\ makes trade-off between statistical accuracy and communication complexity using mini-batch sampling strategies, so as to enjoy low communication costs and accelerated training procedure.


\item Finally, we evaluate \TheName{} using a large-scale real-world user profile dataset and several benchmark datasets for classification. The results show that \TheName{} can compete with state of the art of Gradient Boosting Machines --- LightGBM, XGBoosts and the vanilla re-implementation of LightGBM based on Microsoft SEAL.

\end{itemize}
The rest of the paper is organized as follows. In Section II, we review the gradient-boosting trees based classifiers and the implementation of LightGBM, then we introduce the problem formulation of our work. In Section III, we propose the framework of \TheName{} and present the details of \TheName{} algorithm. In Section IV, we evaluate the proposed algorithms using the real-world user profile dataset and the benchmark datasets. In addition, we compare \TheName{} with baseline centralized algorithms. In Section V, we introduce the related work and present a discussion. Finally, we conclude the paper in Section VI.

\section{Preliminary Studies and Problem Definitions}
In this section, we first present the preliminary studies of the proposed study, then introduce the design goals for the proposed systems as the technical problem definitions.

\subsection{Gradient Boosting and LightGBM}
As an ensemble learning technique, the Gradient Boosting classifier trains and combines multiple weak prediction models, such as decision trees, for better generalization performance~\cite{friedman2001greedy,friedman2002stochastic}.  The key idea of gradient boosting is to consider the procedure of boosting as the optimization over certain cost functions~\cite{breiman1997arcing}. As the result, the gradient descent directions for the loss function minimization can be transformed into the decision trees that were obtained sequentially to improve the classifier. 

Given a training dataset, where each data point $(x,y)\sim\mathcal{D}$, the problem of gradient boosting is to learn a function $\widehat F$ from all possible hypotheses $\mathcal{H}$ while minimizing the expectation of loss over the distribution $\mathcal{D}$, such that
\begin{equation}
    \widehat F = \underset{F\in\mathcal{H}}{\mathrm{argmin}} \underset{(x,y)\sim\mathcal{D}}{\mathbb{E}} L(y, F(x)),
\end{equation}
where $L(y, F(x))$ refers to the prediction loss of $F(x)$ to the label $y$. More specific, the gradient boosting intends to minimize the loss function and obtain $\widehat F$ in a gradient descent way, such that
\begin{equation}
    F_{k+1}(x)\gets F_k(x) + \alpha_k\cdot h_k(x),
\end{equation}
where $F_k(x)$ refers to the learned model in the $k^{th}$ iteration,  $h_k(x)$ refers to the decision tree learned as the descent direction at the $k^{th}$ iteration based on the model already obtained $F_k(x)$ and the training dataset, $\alpha_k$ refers to the learning rate of gradient boosting or namely the weight of $h_k(x)$ in the ensemble of learners, the operator $a+b$ refers to the ensemble of $a$ and $b$ models, and $F_{k+1}(x)$ refers to the results of the $k^{th}$ iteration. More specific, the computation of $h_k(x)$ majorly address $\left(y-F_k(x)\right)$ for $\forall (x,y)\in\mathcal{D}$, i.e.,the error between the model $f_k(x)$ that is already estimated and the label $y$ that corresponds to $x$ in training dataset. Note that in the first iteration, the algorithm starts from $F_1(x)$ which is a vanilla decision tree learned from the dataset. With totally $K$ iteration, the algorithm obtains the final model $\widehat F(x)$ as the $F_{K+1}$.

Recently, gradient boosting classifiers have attracted further attentions from both application and algorithmic perspectives. For example, it has won the KDDCup 2016~\cite{sandulescu2016predicting} and tons of other competition such as Kaggle\footnote{https://medium.com/@gautam.karmakar/xgboost-model-to-win-kaggle-e12b35cd1aad}.  Gradient boosting trees and its variants have been used as a major baselines for a great number of classification/regression tasks with decent results, ranging from genetic data analytic to the click through predictions~\cite{nielsen2016tree}. In terms of algorithm implementation,  XGBoost~\cite{chen2016xgboost} and LightGBM~\cite{ke2017lightgbm}  have been proposed to further improve the performance of gradient boosting trees, where thw two work followed similar gradient boosting mechanisms for the decision trees training while made significant contributions to scalability and efficiency issues.

\subsection{Homomorphic Encryption Models}


To secure the security and privacy during the computation, homomorphic encryption (HE) has been proposed as a set of operations that work on the encrypted data while resulting in the secure ones with encryption. More important the results obtained can be decrypted to match the ``true results'' of the corresponding operations~\cite{gentry2010computing,vaikuntanathan2011computing}. Homomorphic encryption contains multiple types of encryption schemes, such as partially homomorphic encryption (PHE), fully homomorphic encryption (FHE) and pre-fully homomorphic encryption (Pre-FHE), that can perform different classes of computations over encrypted data~\cite{armknecht2015guide}. The progress along these lines of research has been well surveyed in~\cite{halevi2017homomorphic}.

As early as 1978, the tentative idea of building a fully homomorphic encryption scheme was proposed just after the publishing of RSA algorithm~\cite{demillo1978foundations}. Thirty years, Gentry emph{et al.} in 2009  sketched the first fully homomorphic encryption scheme based on the lattice cryptography~\cite{gentry2009fully}. One year later, van Dijk \emph{et al.} presented the second fully homomorphic encryption scheme~\cite{van2010fully} based on Gentry's work, but did not rely on the use of ideal lattices. The second generation of FHE starts from 2011, there were some fundemental techniques developed by Zvika Brakerski \emph{et al.}~\cite{brakerski2014leveled,brakerski2014efficient}, where the homomorphic cryptosystems currently used are stemmed. Thanks to these innovations, the second generation of FHE tends to be much more efficient compared with the first generation, and be applied to a lot of applications.

Later, Gentry \emph{et al.} proposed a new technique for building fully homomorphic encryption schemes, namely GSW, which avoids the use of expensive ``relinearization'' computation in homomorphic multiplication~\cite{gentry2013homomorphic}. Brakerski \emph{et al.} observed that, for certain types of circuits, the GSW cryptosystem features an even slower growth rate of noise, and hence better efficiency and stronger security~\cite{brakerski2014lattice}. 

As the fully homomorphic encryption is computationally expensive, the most of practical secure systems indeed have been implemented with a partially homomorphic encryption fashion~\cite{halevi2017homomorphic}, where only parts of computation are encrypted with homomorphic encryption. In this work, we hope to secure the computation and communication of federated learning through partially homomorphic encryption. Our proposed method uses ciphertexts to protects parts of computations and communications in the gradient boosting trees learning.

\subsection{Problems and Overall Design Goals}
In this work, we intend to design a novel federated gradient boosting trees classifier that can learned from view separated data in a distributed manner while avoiding the leakage of data privacy and security.

\textbf{The Federated Learning Problem -} Suppose two training datasets $\mathcal{D}_1$ and $\mathcal{D}_2$ are owned by two parties $\mathbb{A}$ and $\mathbb{B}$ respectively, who hope to collaboratively learn one mode but don't trust each other. The schemes of the two datasets are $\mathcal{D}_1=(I; X; Y)$ and $\mathcal{D}_2=(I;X)$, where
\begin{itemize}
    \item $I(\mathcal{D}_1)$ and $I(\mathcal{D}_2)$ refer to the identity sets of samples in the two datasets respectively. When $I(\mathcal{D}_1)\cap I(\mathcal{D}_2)\neq\emptyset$, it indicates that the two datasets share a subset of samples but with different views (i.e., features);
    \item $X(\mathcal{D}_1)$ and $X(\mathcal{D}_2)$ refer to the feature sets of samples in the two datasets respectively. More specific, we denote $X_i(\mathcal{D}_1)$ as the set of features of the $i^{th}$ sample in $\mathcal{D}_1$ and further denote $X_{i,j}(\mathcal{D}_1)$ as the $j^{th}$ feature of the $i^{th}$ sample in $\mathcal{D}_1$;
    
    \item $Y(\mathcal{D}_1)$ refers the label set of the data samples in $\mathcal{D}_1$. More specific, $\forall y\in Y(\mathcal{D}_1)$ there has $y\in\mathbb{R}^{|\mathcal{C}|}$ and $\mathcal{C}$ refers to the set of classes. As was mentioned, in the settings discussed in this paper, only one part would monopoly the label information.
\end{itemize}
With above settings, our federated learning problem consists of two parts as follow.

\textbf{Training - } There needs to propose an dual-party learning algorithm with secure communication and computation schemes that can train tree-based Gradient Boosting Machines based on $\mathcal{D}_1$ and $\mathcal{D}_2$ with respect to the following restrictions:
\begin{itemize}
    \item \textbf{Training Set Identity Protection}: $I(\mathcal{D}_1)\backslash I(\mathcal{D}_2)$ would not be obtained by the party $\mathbb{B}$, while the information about $I(\mathcal{D}_2)\backslash I(\mathcal{D}_1)$ would be prohibited from $\mathbb{A}$;
    
    \item \textbf{Training Set Feature Security}: The inference procedure needs to avoid the leakage of $X(\mathcal{D}_1)$ and $Y(\mathcal{D}_1)$ to $\mathbb{B}$, and the  $X(\mathcal{D}_2)$ to the party $\mathbb{A}$.
\end{itemize}

\textbf{Testing - } Given two testing datasets $\mathcal{D'}_1=(I;X)$ and $\mathcal{D'}_2=(I;X)$ owned by the two parties $\mathbb{A}$ and $\mathbb{B}$ respectively, where $I(\mathcal{D'}_1)\cap I(\mathcal{D'}_2)\neq\emptyset$. There needs an online inference algorithm, where the party $\mathbb{A}$ can initialize the inference procedure using the identity of the sample for prediction (i.e., $i'\in I(\mathcal{D'}_1)\cap I(\mathcal{D'}_2)$). The party $\mathbb{A}$ can obtains the prediction result of the sample (i.e., $y_{i'}$) through the secure inference procedure, with respect to the following restrictions:
\begin{itemize}
    \item \textbf{Testing Set Identity Protection}:  $I(\mathcal{D'}_1)\backslash I(\mathcal{D'}_2)$ would not be obtained by the party $\mathbb{B}$, while the information about $I(\mathcal{D'}_2)\backslash I(\mathcal{D'}_1)$ would be prohibited from $\mathbb{A}$
    
    \item \textbf{Testing Set Feature Security}: The inference procedure needs to avoid the leakage of $X(\mathcal{D'}_1)$ and $y_{i'}$ to $\mathbb{B}$, and $X(\mathcal{D'}_2)$ to the party $\mathbb{A}$.
\end{itemize}
In our research, we intend to design PHE-based encryption schemes to protect the training and inference procedures (derived from LightGBM~\cite{ke2017lightgbm}) and meet above security goals.


\begin{figure*}
    \centering
    \includegraphics[width=0.95\textwidth]{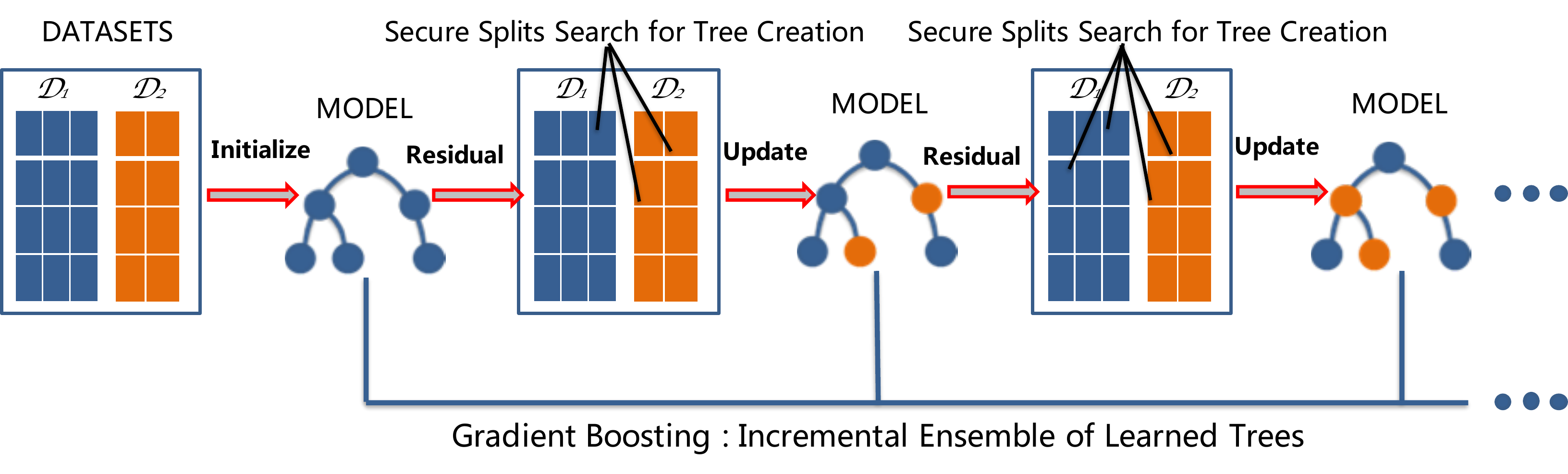}
    \caption{The Training Procedure of \TheName}
    \label{fig:train_proc}
\end{figure*}

\section{Frameworks and Algorithms Design}
In this section, we present the framework design of \TheName\ with key algorithms used.

\subsection{Overall Framework Design}
The overall framework of \TheName\ consists of two parts --- training and inference, where, given the distributed datasets, the training procedure obtains the distributed parameter models for the tree classifiers of \TheName\ and the inference procedure predicts the labels using the indices of samples. 

\subsubsection{Statistically Accelerated Training Procedure} Given the training datasets $\mathcal{D}_1$ and $\mathcal{D}_2$ distributed in the two parties, as shown in Figure~\ref{fig:train_proc}, the training procedure learns the ensemble of decision trees for Gradient Boosting Machines with distributed parameters in a secure and statistical efficient way. More specific, the training procedure incorporates an accelerated iterative process with a specific initialization as follow
\begin{itemize}
    \item \textbf{Initialization - } The owner of $\mathcal{D}_1$ can invoke to initialize the whole training procedure. First of all, \TheName\ performs \emph{secure join operation} to align the shared samples stored in $\mathcal{D}_1$ and $\mathcal{D}_2$
    through matching $I(\mathcal{D}_1)$ and $I(\mathcal{D}_2)$ under \emph{Partial Homomorphic Encryption (PHE)} settings. Later, based on the data in $\mathcal{D}_1$ including both features $X(\mathcal{D}_1)$ and labels $Y(\mathcal{D}_1)$ , \TheName\ learns a decision tree $F_0$ as the base model, which only uses features in $X(\mathcal{D}_1)$, for initialization. Please see also in Section III.B.1 for the detailed design and implementation of \emph{Secure Join Operation} for sample alignment based on PHE.
\end{itemize}

With the model initialized, \TheName\ takes a statistically accelerated iterative process for GBM training, where each iteration uses \emph{mini-batch sampling to reduce the computational/communication costs}~\cite{friedman2002stochastic}. Specifially, each iteration (e.g., the $k^{th}$ iteration and $k\geq 0$) consists of three steps:
\begin{itemize}   
    \item \textbf{Batched Secure Inference - } Given the shared samples in $\mathcal{D}_1\cap\mathcal{D}_2$, \TheName\ first randomly selects a subset of samples $\mathcal{B}_k\subseteq (I(\mathcal{D}_1)\cap I(\mathcal{D}_2))$, where $|\mathcal{B}_k|=b$ and $b$ refers to the batch size. With the model already estimated, denoted as $F_{k}$, \TheName\ then obtains the \emph{``soft prediction''} results of all samples in $\mathcal{B}_k$ through the secure inference under PHE settings. Such that
    \begin{equation}
        y_i^{k}\gets F_{k}(X_i(\mathcal{D}_1);X_i(\mathcal{D}_2)),\ \ \forall i \in \mathcal{B}_k,
    \end{equation}
where $F_{k}(X_i(\mathcal{D}_1);X_i(\mathcal{D}_2))$ refers to the inference result based on the features from both datasets. Please see also in Section III.A.2 for the PHE-based implementation of the inference procedure. 
    \item \textbf{Residual Error Estimation - }  As was mentioned, both labels and soft prediction results are a $|\mathcal{C|}$-dimensional vector, where $|\mathcal{C}|$ refers to the number of classes.
    Then, \TheName\ estimates the residual errors of the current model using the \emph{cross-entropy} as follow
        \begin{equation}
        \varepsilon_i^{k}\gets \mathrm{H}(y_i^{k})+ \mathrm{D_{KL}}\left( y_i^{k}||Y_i(\mathcal{D}_1)\right),\ \ \forall i \in \mathcal{B}_k.
    \end{equation}
Note that to secure the security of labels, $y_i^{k}$ and $\varepsilon_i^{k}$ for $\forall i\in \mathcal{B}_k$ are all stored in the owner of $\mathcal{D}_1$.
    \item \textbf{Secure Tree Creation and Ensemble - } Given the residual error estimated $\varepsilon_i^{k},\ \forall i \in\mathcal{D}_1\cap\mathcal{D}_2$, \TheName\ boosts the learned model $F_{k}$ through creating a new decision tree $h_k$ that fits the residual errors $\varepsilon_i^{k}$ using the features of both datasets $X(\mathcal{D}_1)\cup X(\mathcal{D}_2)$ in an additive manner. \TheName\ then ensembles $h_k$ with the current model $F_k$ and obtains the new model $F_{k+1}$ through gradient boosting~\cite{friedman2001greedy}. As was mentioned in Eq.~(2), a specific learning rate $\alpha$ has been given as the weight for model ensembling.  Please see also in Section III.B.2 for the detailed design and implementation of \emph{Secure Splits Search Operation} for decision tree creation based on PHE.
\end{itemize}

\subsubsection{Inference Procedure via Execution Flow Compilation} 
\begin{figure}
    \centering
    \includegraphics[width=0.48\textwidth]{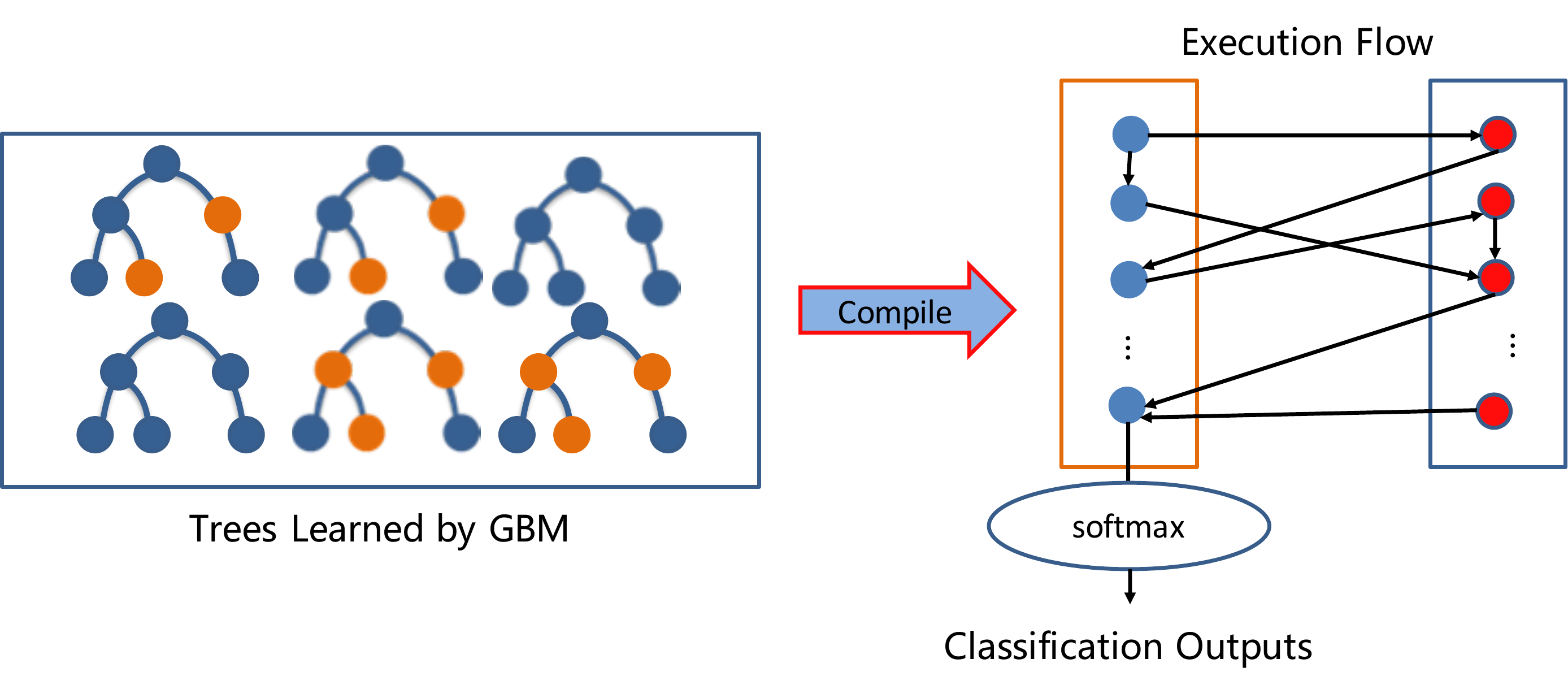}
    \caption{Execution Flow Compilation for the Inference Procedure}
    \label{fig:infer_proc}
\end{figure}

Given the model learned after $K$ iterations, denoted as $F_{K}$, the inference component first complies all these trees into distributed secure execution flow, where the nodes in every decision trees are assigned to the corresponding parties respectively. As shown in Figure~\ref{fig:infer_proc}, all \emph{communications}, \emph{computation} and \emph{binary comparisons} are protected through SEAL-based Homomorphic Encryption schemes. With the secure distributed execution flow, given a index of sample e.g., $i'\in I(\mathcal{D'}_1)\cap I(\mathcal{D'}_2)$, \TheName\ runs the inference procedure over the execution flow. Please see also in Section III.B.3 for he design and implementation of \emph{PHE-based Binary Comparison Operator}. Note that in our research, we assume the party $\mathbb{B}$ has no way to access the labels of training and testing samples, while securing the monopoly of the label information at the party $\mathbb{A}$ side. To protect the label information through the inference, the result of PHE-based Binary Comparison Operator (i.e., true or false) has been secured and cannot be deciphered by the party $\mathbb{A}$.

Furthermore, for the sample with index that is contained in $I(\mathcal{D'}_1)$ only, i.e., $i'\in I(\mathcal{D'}_1)\backslash I(\mathcal{D'}_2)$, \TheName\ would first learn a comprehensive GBM (LightGBM classifier) based on the dataset $\mathcal{D}_1$ using the features $X(\mathcal{D}_1)$ and labels $Y(\mathcal{D}_1)$ only. With such model, \TheName\ makes prediction for the samples in $I(\mathcal{D'}_1)\backslash I(\mathcal{D'}_2)$ all using the features in $X(\mathcal{D'}_1)$.

Please note that the overall framework of \TheName\ is derived from the vanilla implementation of LightGBM~\cite{ke2017lightgbm}, while most of calculation and optimization to gradient boosting trees~\cite{friedman2001greedy} has been preserved with coverage of partial homomorphic encryption. 

\subsection{Key Algorithms}
Here, we present the detailed design of several key algorithms.

\subsubsection{Secure Join for Sample Alignment} To align the samples with identical indices across the two index sets $I(\mathcal{D}_1)$ and  $I(\mathcal{D}_2)$ for training (and  $I(\mathcal{D'}_1)$ and  $I(\mathcal{D'}_2)$ for inference), \TheName\ intends to obtain the intersection between the two index sets in a private and secure manner. Specifically, we adopt the private sets intersection algorithms proposed in~\cite{pinkas2014faster,pinkas2018scalable} to achieve the goal. The security and  privacy eniforcement of proposed component highly relies on the techniques called Obvious Transfer Extension (OT Extension), which supports fast and private data transfer for small payloads with limited overhead~\cite{keller2015actively}. The use of OT Extension can avoid the use of time-consuming error correcting code but instead accelerate the secure data transmission through leveraging a pseudo-random code. We also tried other OT extension based private sets intersection algorithm, such as the one using Bloom Filter~\cite{dong2013private}. The speed and scalability is not as good as~\cite{pinkas2014faster,pinkas2018scalable}.

\subsubsection{Secure Splits Search for Tree Creation} For each round of iteration (e.g., the $k^{th}$ iteration), \TheName\ needs to create a decision tree with the size $t$ (here $t$ refers to the number of learned split nodes in the tree) to fit the ``residual error'' of the model that is already estimated $F_k$, as as to enable the gradient boosting mechanism. Specifically, we adopt a leaf-wise tree growth mechanism that was derived from LightGBM~\cite{ke2017lightgbm} to learn the tree, where \TheName\ vertically grows the tree using totally $t$ rounds of computation/communication, and it always picks up the leaf node with maximal ``residual error'' reduction to grow. 

In each round of computation for the decision tree creation, for the party $\mathbb{A}$ owning features $X(\mathcal{D}_1)$ and labels $Y(\mathcal{D}_1)$, \TheName\ searches new splits using the raw data. Similar to vanilla LightGBM, \TheName\ selects the ``best'' split with the maximal residual error reduction for the samples in the mini-batch $\mathcal{B}_k$ as the a candidate of the split at the party $\mathbb{A}$ side. This candidate split would be compared to the ``best'' splits from the party $\mathbb{B}$ as the final split selection for this round.

On the other hand, for the party $\mathbb{B}$ only occupying features $X(\mathcal{D}_2)$, \TheName\ first propose a set of potential splits in $X(\mathcal{D}_2)$ (in a random or unsupervised manner), and sends the potential classification results of the samples in $\mathcal{B}_k$ using every proposed split to the party $\mathbb{A}$. Note that the potential classification results have been formed into multiple sets of samples (or sample index), which have been categorized according to their results. Such sets have been encrypted as private sets to protect the privacy of label information from the party $\mathbb{B}$. Certain secure domain isolation has been used to protect the splits~\cite{liu2015thwarting}.

Then, at the party $\mathbb{A}$ side, \TheName\ estimates the residual errors of each split proposed by the party $\mathbb{B}$ using their potential classification results. Specifically, \TheName\ leverage the aforementioned private set intersection algorithm to estimate the overlap between the sample sets categorized using potential classification results and the true labels, in order to obtain the prediction results and estimate the accuracy~\cite{pinkas2014faster,pinkas2018scalable}. Finally, \TheName\ selects the split (the best splits from $\mathbb{A}$ versus $\mathbb{B}$) that can further lower residual error as the split in this round and ``adds'' the split to the decision tree.

To further secure the privacy of label information, the splits at the party $\mathbb{B}$ are deployed in an isolated domain, while the party $\mathbb{B}$ cannot obtain the decision making result of splits. Please refer the section in below for the implementation of the binary comparison.

\subsubsection{Secure Binary Comparison for Decision Making} As was mentioned, \TheName\ operates an isolated domain over the machines at the party $\mathbb{B}$, where the computation and comparison criterion for decision making are all stored in the isolated domain, which is trusted by both parties. To further secure the label information and prediction results during inference, \TheName\ uses the public keys generated the party $\mathbb{A}$ to encrypted the decision making results from $\mathbb{B}$, while the public keys keep being updated per inference task.

\subsection{Discussion and Remarks}
In this section, we intends to justify the proposed framework and algorithms from  costs and learning perspectives.

\textbf{Communication Costs - } In \TheName, we replaced the gradient boosting mechanism used by GBM with stochastic gradient boosting~\cite{friedman2002stochastic}, in order to accelerate the learning procedure through lowering the computational/communication costs per iteration. Let denote $\mathcal{N} = |I(\mathcal{D}_1)\cap I(\mathcal{D}_2)|$ as the total number of aligned samples shared by $\mathcal{D}_1$ and $\mathcal{D}_2$, while the size of batch for each iteration is defined as $b=|\mathcal{B}_k|$.

For each iteration of GBM and \TheName, there needs to create a $t$-sized decision tree after $t$ rounds of communication between the two parities. For each round of such communication, GBM and \TheName\ need to exchange data with payloads size of $\mathcal{O}(\mathcal{N}^2)$ and $\mathcal{O}(b^2)$, respectively. In this way, the cost of communication per iteration should be $\mathcal{O}(t\cdot\mathcal{N}^2)$ and $\mathcal{O}(t\cdot b^2)$ for GBM and \TheName.

\textbf{Statistical Acceleration - } To simplify our analysis on the statistical performance, we make an mild assumption that considers the learning procedure of LightGBM and \TheName\ as the gradient descent (GD) and stochastic gradient descent (SGD) based loss minimization over certain convex loss~\cite{friedman2001greedy,mason2000boosting,friedman2002stochastic}. Under mild convexity and smoothness conditions, GD and SGD would converge to the minimum of the loss functions at the error convergence rate\cite{shapiro1996convergence,shalev2009stochastic,shamir2013stochastic} of $\mathcal{O}(1/k)$ and $\mathcal{O}(1/\sqrt{k})$ respectively, where $k$ denotes the number of iterations. More discussion can be found in~\cite{mason2000boosting}.
While the costs of each iteration are $\mathcal{O}(t\cdot\mathcal{N}^2)$ and $\mathcal{O}(t\cdot b^2)$ respectively, we can roughly conclude a trade-off exists between the statistical performance and communication complexity for \TheName\ Training.


\section{Experiments and Results}
In this section, we mainly report the experiments that evaluate \TheName, and compares the performance of \TheName\ with other baseline methods including vanilla LightGBM, XGBoost, and other downstream classifiers.

\begin{table*}[]
    \centering
        \caption{Overall Classification AUC (\%) Comparison (N/A: During the experiments, LightGBM reported failure to train the model due as the features of the given datasets are too sparse to learn.)}
        {
    \begin{tabular}{r|cc||cc||cc} \hline
         & \multicolumn{2}{c}{Sparse} & \multicolumn{2}{c}{Adult} & \multicolumn{2}{c}{Phishing}  \\ \hline
       Methods & Training & Testing & Training & 
       Testing & Training & Testing \\ \hline
       \TheName &  93.227  &  66.220  &  92.465  & 90.080  &  62.855  & 61.823  \\ \hline
        \multicolumn{7}{c}{Using the Aggregated Datasets from $\mathbb{A}$ and $\mathbb{B}$}\\\hline
       LightGBM-($\mathbb{A}$,$\mathbb{B}$) &  96.102  &  68.528  & 92.199  & 90.145  &  67.994  &  63.430  \\  
       XGBoost-($\mathbb{A}$,$\mathbb{B}$)  &  93.120  &  67.220  & 91.830  &  89.340 &  67.090  &  61.990  \\  
       LIBSVM-Linear-($\mathbb{A}$,$\mathbb{B}$)  &  73.490  &  64.560  &  58.641  &  59.280  &  50.073  &  50.980  \\ 
       LIBSVM-RBF-($\mathbb{A}$,$\mathbb{B}$)     &  79.850  &  63.210  &  75.549  &  72.060  &  52.789  &  47.479  \\ \hline 
        
        \multicolumn{7}{c}{Using the Single Dataset at $\mathbb{A}$}\\\hline 
       LightGBM-$\mathbb{A}$      &  N/A*     &  N/A*     &  89.849  &  88.052  &  64.693  &  59.743  \\  
       XGBoost-$\mathbb{A}$       &  65.170  &  57.370  &  89.490  &  87.620  &  64.070  &  59.740  \\  
       LIBSVM-Linear-$\mathbb{A}$ &  52.360  &  50.675  &  66.293  &  34.347  &  50.007  &  50.489  \\ 
       LIBSVM-RBF-$\mathbb{A}$    &  56.740  &  52.380  &  72.909  &  55.076  &  50.248  &  50.306  \\ \hline 
       
       \multicolumn{7}{c}{Using the Datasets that aggregate Features from  $\mathbb{B}$ and Labels from $\mathbb{A}$ (Not exist in the real case)}\\\hline 
       LightGBM-$\mathbb{B}$*      &  96.102  &  68.528  &  85.708  &  84.587  &  62.396  &  58.929  \\  
       XGBoost-$\mathbb{B}$*       &  93.190  &  67.390  &  85.700  &  85.410  &  61.720  &  58.420  \\  
       LIBSVM-Linear-$\mathbb{B}$* &  67.480  &  60.990  &  46.527  &  46.840  &  50.627  &  48.567  \\ 
       LIBSVM-RBF-$\mathbb{B}$*    &  78.230  &  64.880  &  56.927  &  74.987  &  50.336  &  50.415  \\ \hline 
    \end{tabular}}
    \label{tab:overall}
\end{table*}

\subsection{Experimental Setups}
In this section, we present the dataasets, baseline algorithms as well as tbe experimental settings of our evaluation study.

\subsubsection{Datasets} In our study, we intend to evaluate \TheName\ using three datasets as follow.
\begin{itemize}
    \item \textbf{\emph{Sparse - }} This is a private dataset consisting of 11,371 users' de-anonymized financial records data, where each sample with 8,922 \emph{extremely sparse} features and a binary label. These features are separately owned by two parties --- the bank is with 5,000 features and the real estate loaner owns the rest 3,922 features, while the bank owns the label information about bankruptcy. The goal of this dataset is to predict the bankruptcy of a user, incorporating their sparse features distributed in the two parties. As the dataset is quite large, we set the mini-batch size $b$ as the 1\% of the overall training set. Note that the labels in Sparse are extremely imbalanced, while most samples are negative.
    
    \item \textbf{\emph{Adult - }} This an open-access dataset consisting of 27,739 web pages' information, where each web page is with 123 features and 1 binary label (whether the web page contains adult contents). We randomly split the features into two sets, each of which is with 61 and 62 features respectively. As this dataset is quite small, we use the whole dataset for each iteration i.e., $b=100\%$ of the overall training set. 
    
    \item \textbf{\emph{Phishing - }} This an open-access dataset consisting of 92,651 web pages' information, where each web page is with 116 features and 1 binary label (whether the web page contains phishing risk). We randomly split the features into two sets, each of which is with 58 features respectively.  As this dataset is comparatively large, we use the the mini-batch with $b=10\%$ of the overall training set. 
\end{itemize}

\subsubsection{Baseline Algorithms with Settings} To understand the advantage of \TheName, we compare \TheName\ with the baseline algorithms in below.
\begin{itemize}
    \item \textbf{\emph{LightGBM - }} We consider the vanilla implementation of LightGBM as a key baseline algorithm, where we include two settings LightGBM-$\mathbb{A}$ and LightGBM-$(\mathbb{A},\mathbb{B})$. LightGBM-$\mathbb{A}$ refers to the LightGBM that is trained using the features and labels in the dataset $\mathbb{A}$ only, while LightGBM-$(\mathbb{A},\mathbb{B})$ refers to the vanilla distributed LightGBM that is trained using the both datasets in $\mathbb{A}$ and $\mathbb{A}$, without encryption protection. Finally, we also include a baseline LightGBM-$\mathbb{B}*$ that might not exist in the real-world settings --- LightGBM-$\mathbb{B}*$ aggregates the features from the party $\mathbb{B}$ and label information from $\mathbb{A}$ as the training dataset. The comparison to LightGBM-$\mathbb{A}$ and LightGBM-$\mathbb{B}*$ would show the information gain of federated learning beyond the model that was trained using any single party.

    \item \textbf{\emph{XGBoost - }} Following the above settings, we include two settings XGBoost-$\mathbb{A}$ and XGBoost-$(\mathbb{A},\mathbb{B})$. XGBoost-$\mathbb{A}$ refers to the vanilla XGBoost that was trained using the features and labels in the dataset $\mathbb{A}$ only, while XGBoost-$(\mathbb{A},\mathbb{B})$ refers to the vanilla XGBoost trained through aggregating the both datasets from $\mathbb{A}$ and $\mathbb{B}$, in a centralized manner. Similarly,  XGBoost-$\mathbb{B}*$ that was trained through aggregating the features from the party $\mathbb{B}$ and label information from $\mathbb{A}$ was given a baseline to demostrate the information gain of collaboration.

    \item \textbf{\emph{LIBSVM - }} Following the same settings, we include two settings LIBSVM-$\mathbb{A}$ and LIBSVM-$(\mathbb{A},\mathbb{B})$. LIBSVM-$\mathbb{A}$ refers to the vanilla LIBSVM that is trained using the features and labels in the dataset $\mathbb{A}$ only, while LIBSVM-$(\mathbb{A},\mathbb{B})$ refers to the vanilla XGBoost that is trained using the both datasets in $\mathbb{A}$ and $\mathbb{A}$, in a centralized manner. Similarly,  SVM-$\mathbb{B}*$ that was trained through aggregating the features from the party $\mathbb{B}$ and label information from $\mathbb{A}$ was given a baseline. More specific, the LIBSVM algorithms with RBF kernel and linear SVM are used here.

\end{itemize}
Note that in all experiments, 80\% samples are used for training and the rest 20\% samples are remained for testing. The training and testing sets are randomly selected for 5-folder cross validation. The default learning rate for LightGBM, XGBoost and \TheName\ are all set to 0.1.

\begin{figure*}
    \centering
    \subfloat[$t=5$ on Sparse]{\includegraphics[width=0.3\textwidth]{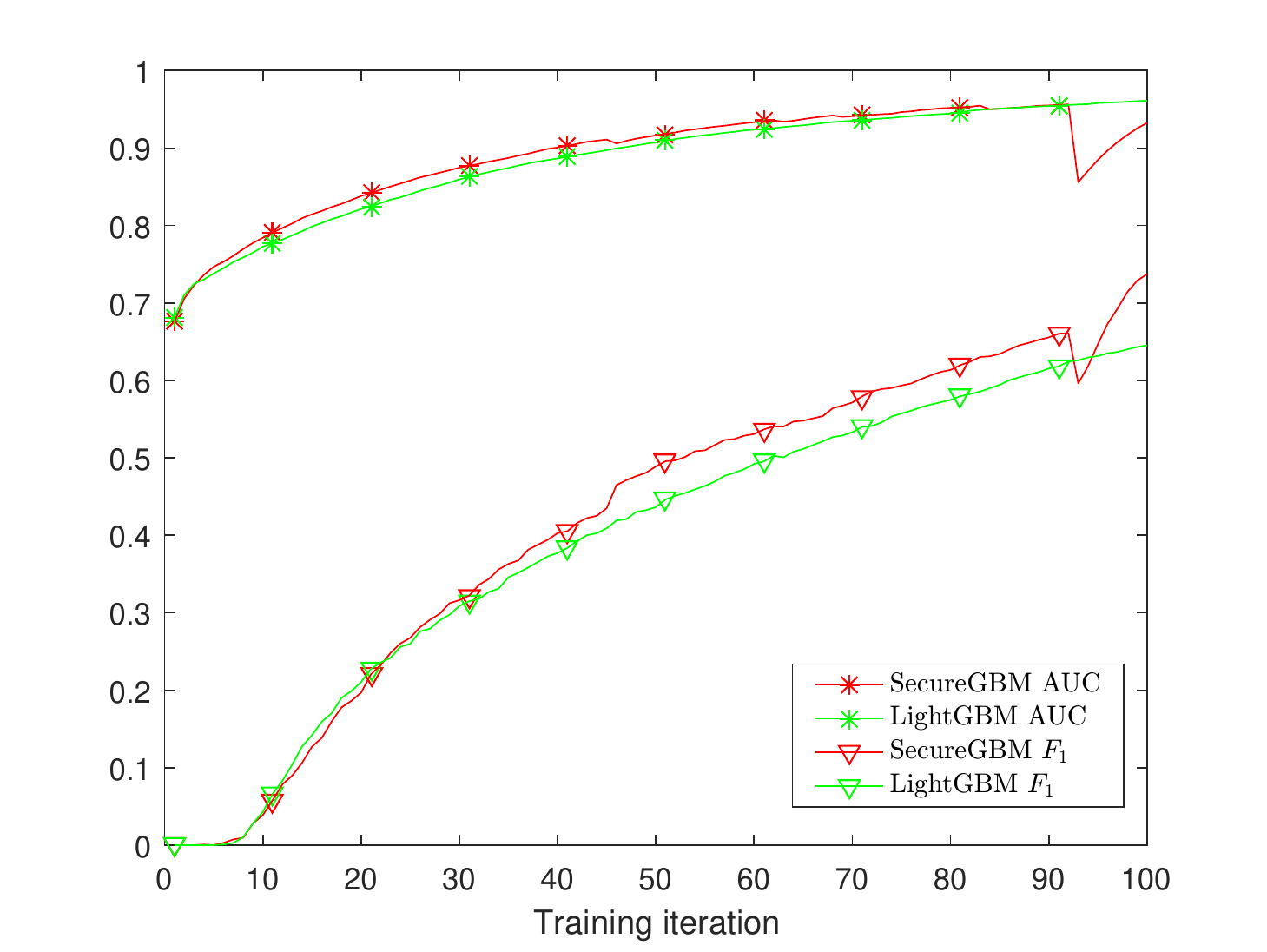}}
    \subfloat[$t=4$ on Sparse]{\includegraphics[width=0.3\textwidth]{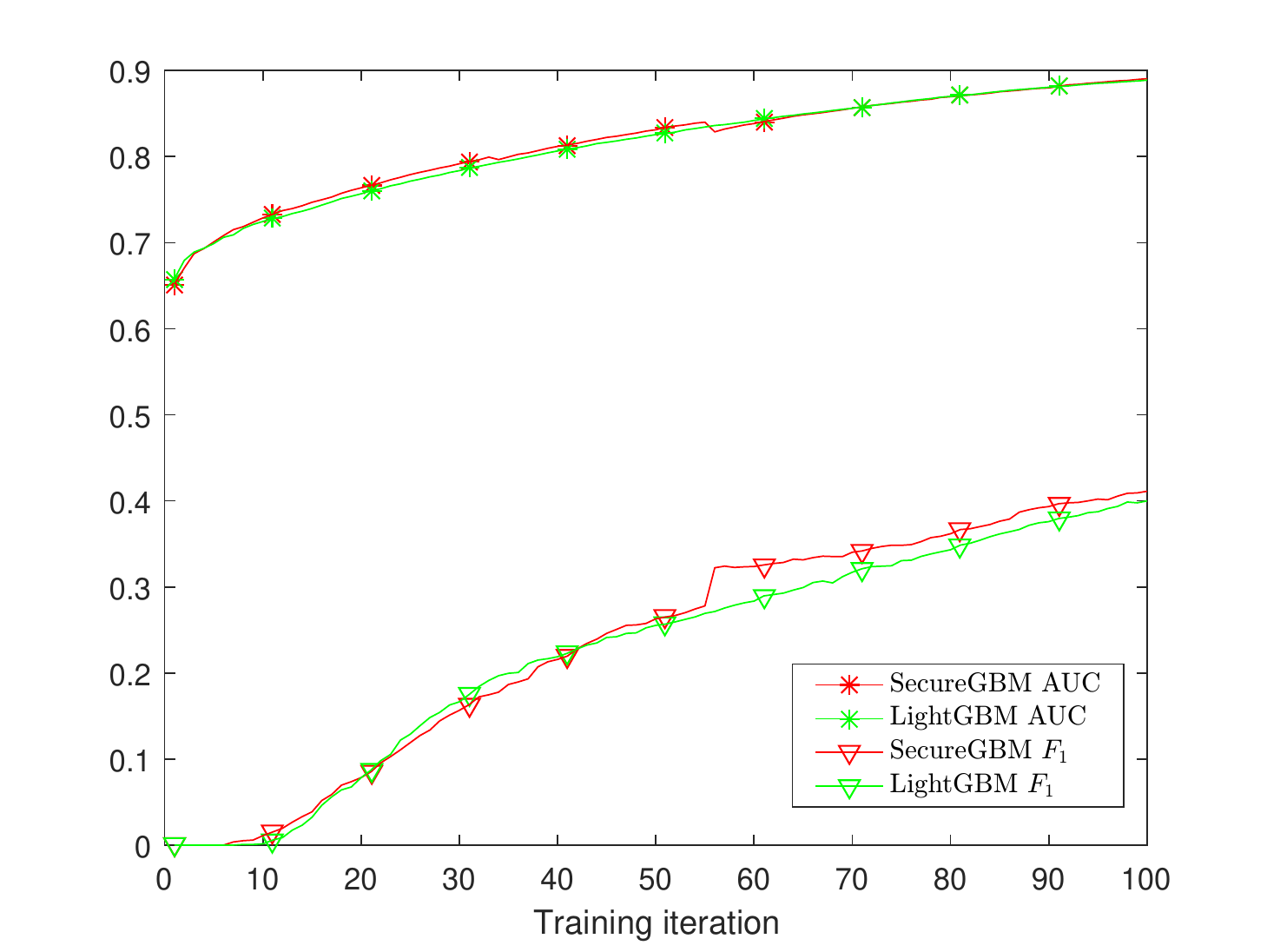}}
    \subfloat[$t=3$ on Sparse]{\includegraphics[width=0.3\textwidth]{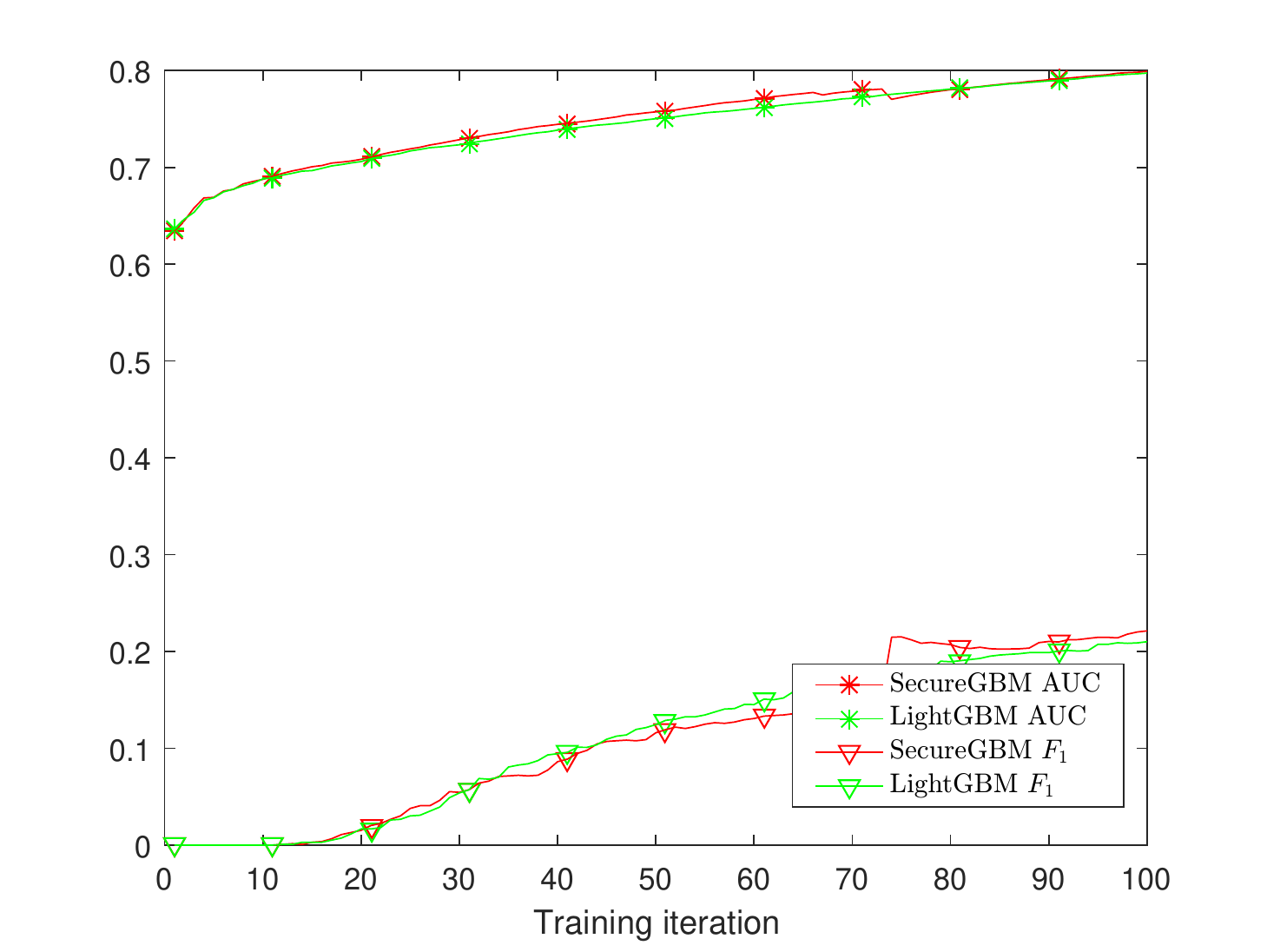}}\\
    \subfloat[$t=5$ on Adult]{\includegraphics[width=0.3\textwidth]{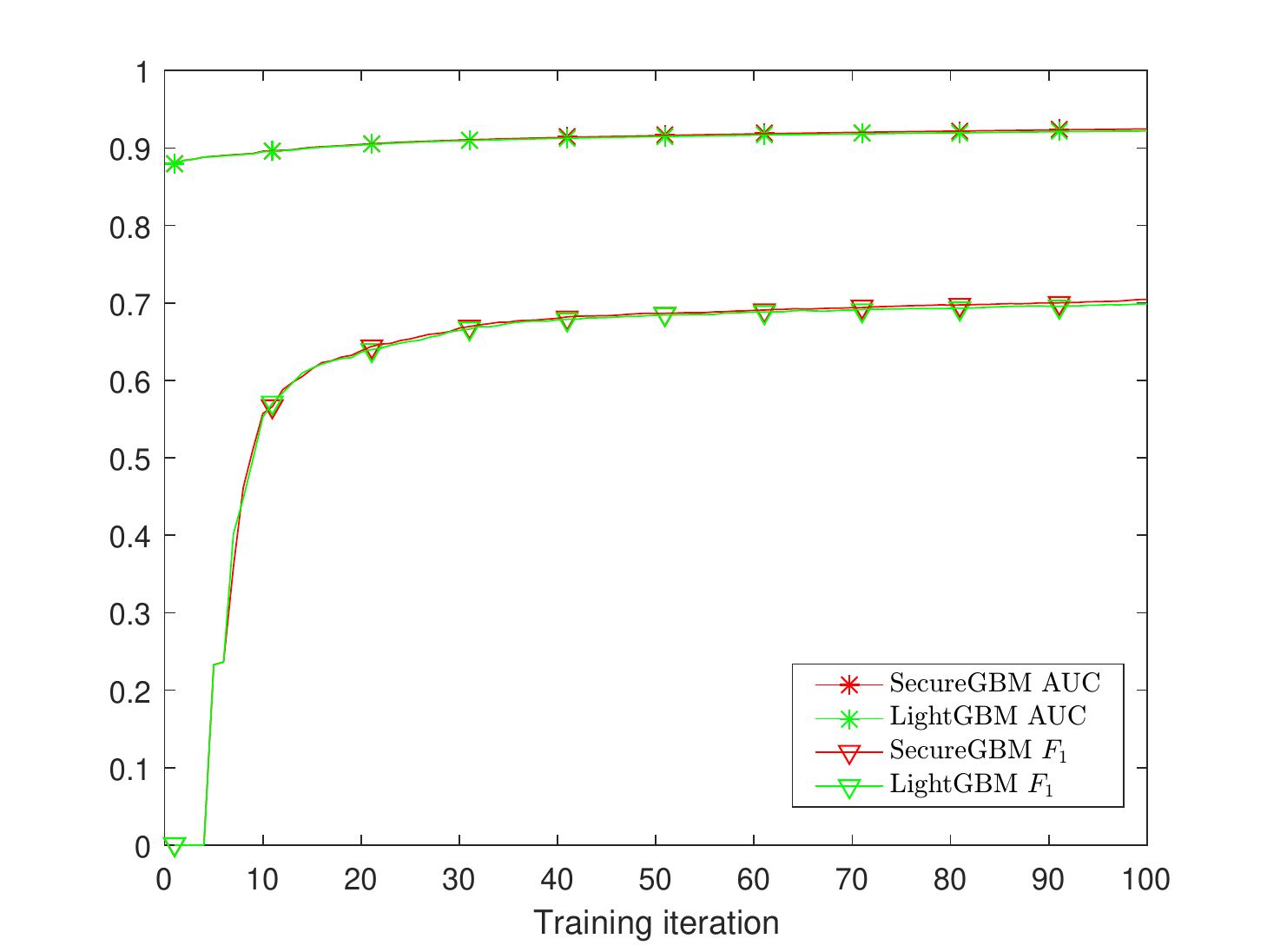}}
    \subfloat[$t=4$ on Adult]{\includegraphics[width=0.3\textwidth]{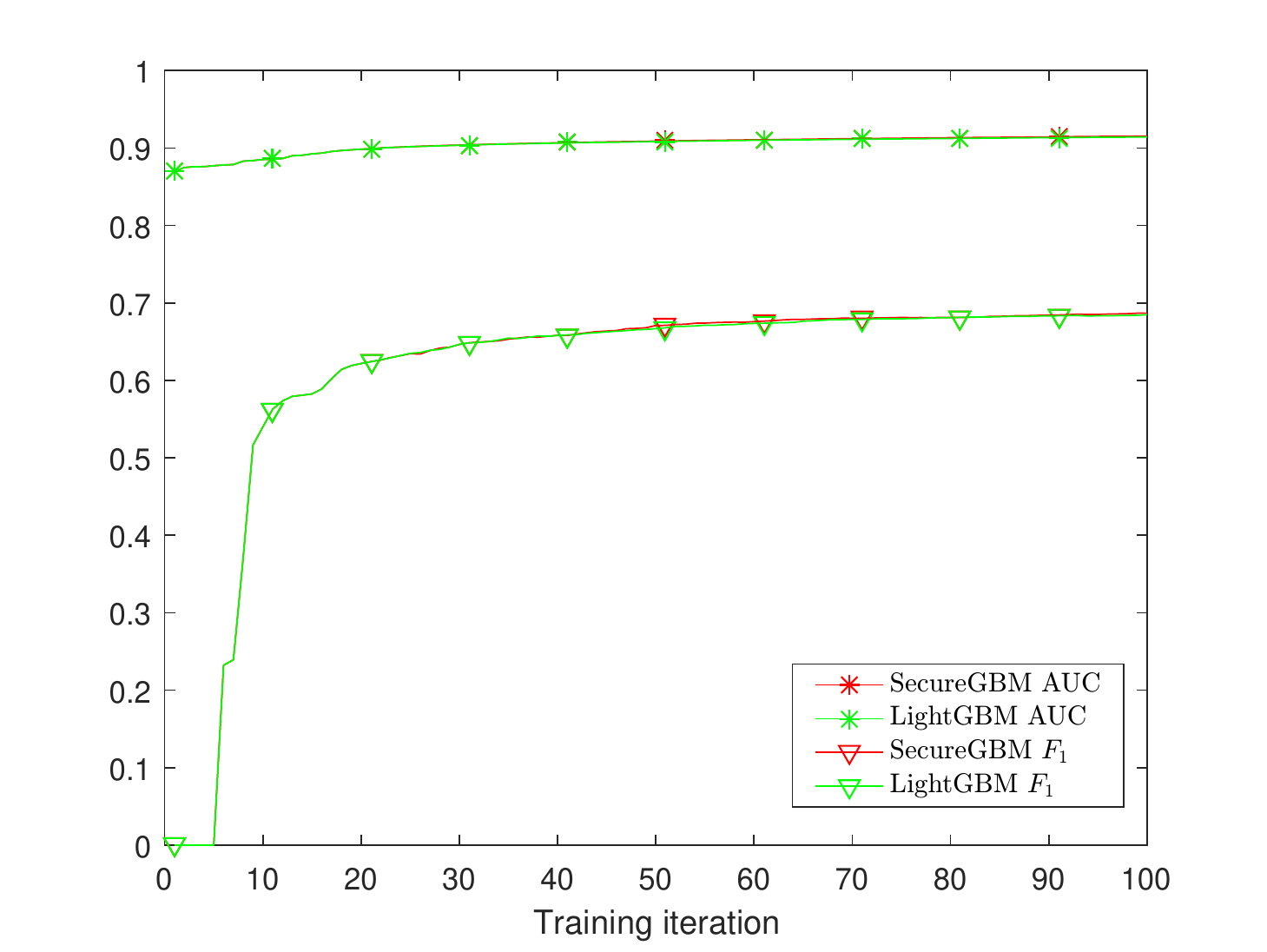}}
    \subfloat[$t=3$ on Adult]{\includegraphics[width=0.3\textwidth]{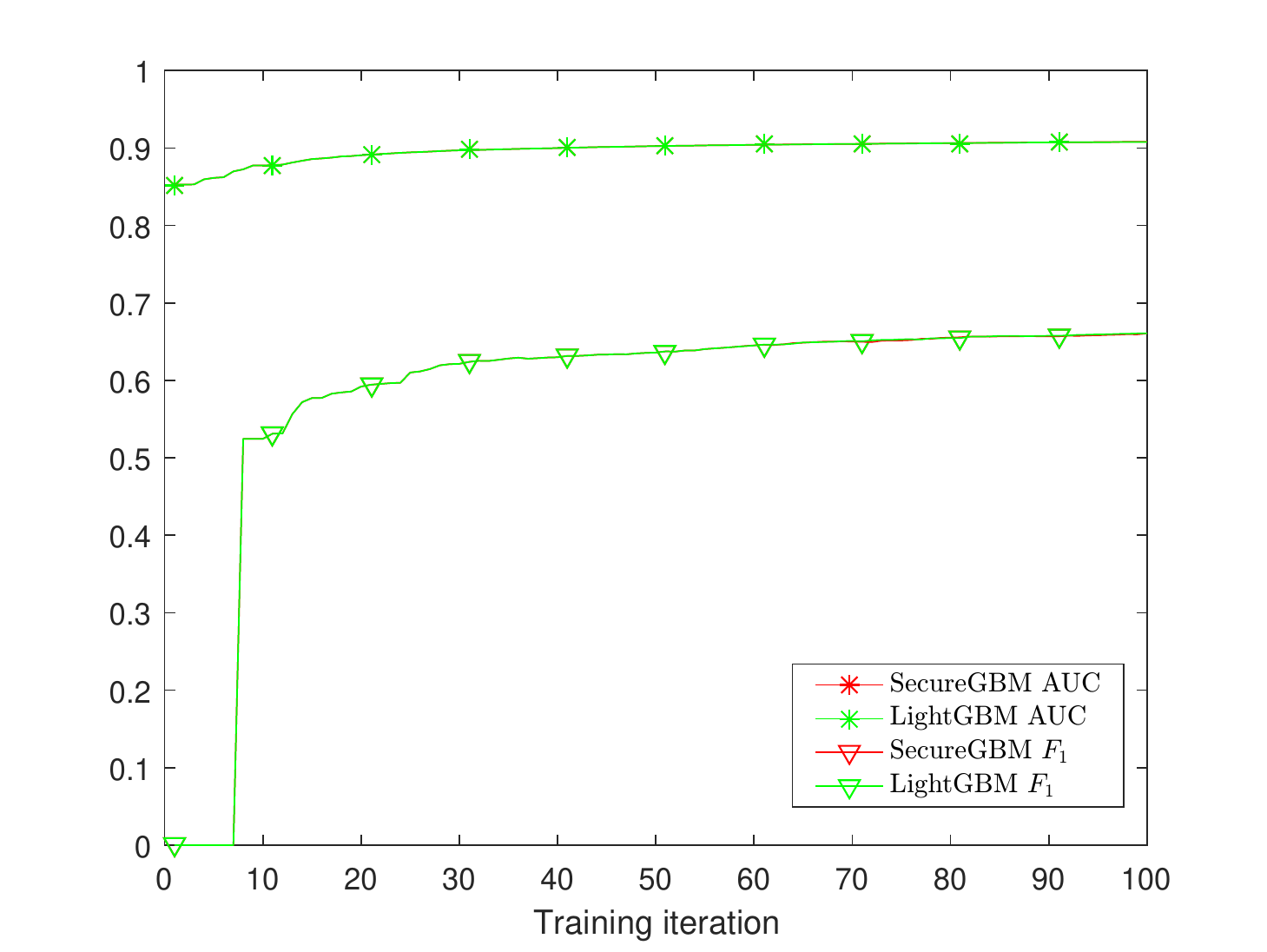}}\\
    \subfloat[$t=5$ on Phishing]{\includegraphics[width=0.3\textwidth]{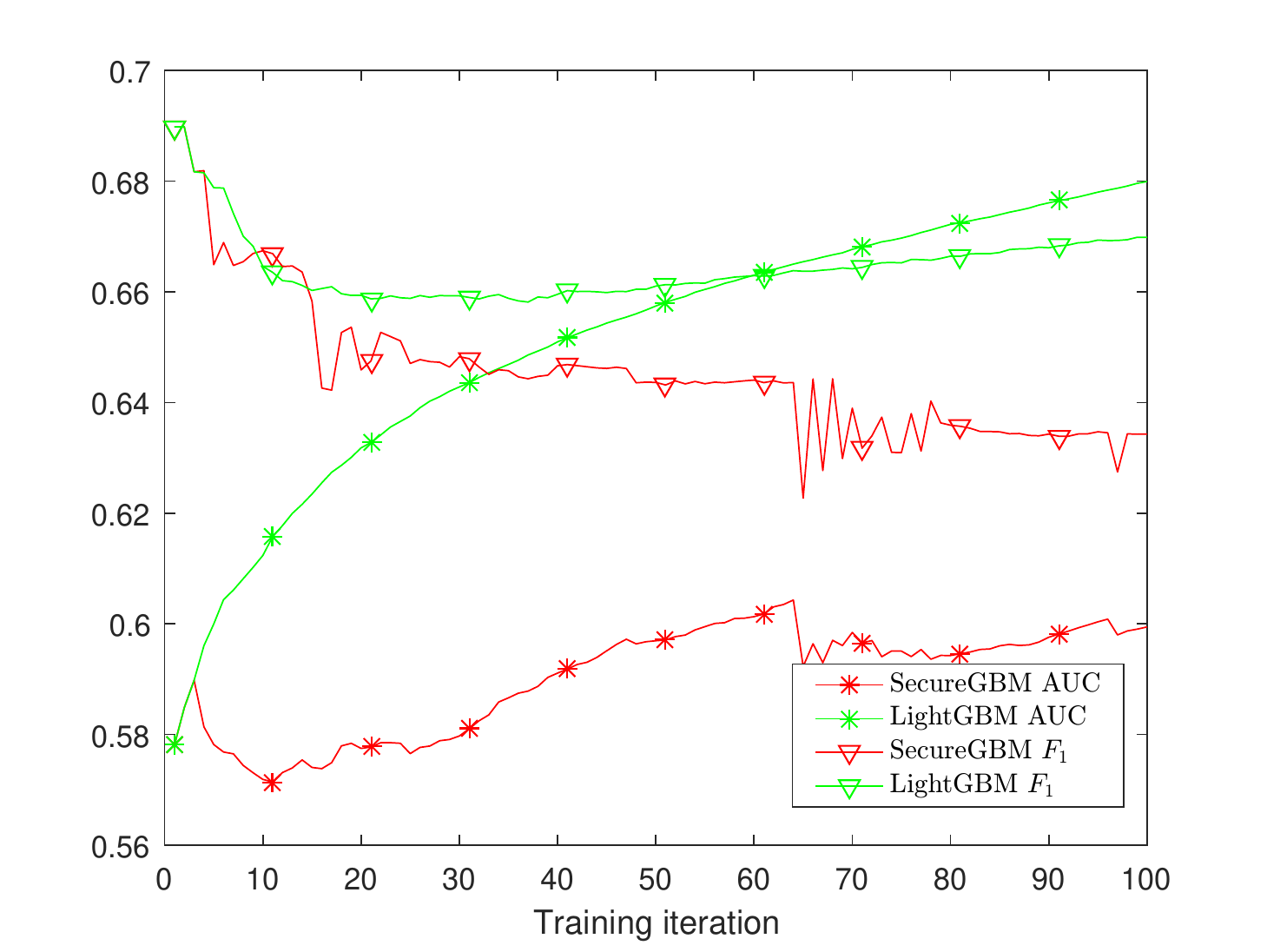}}
    \subfloat[$t=4$ on Phishing]{\includegraphics[width=0.3\textwidth]{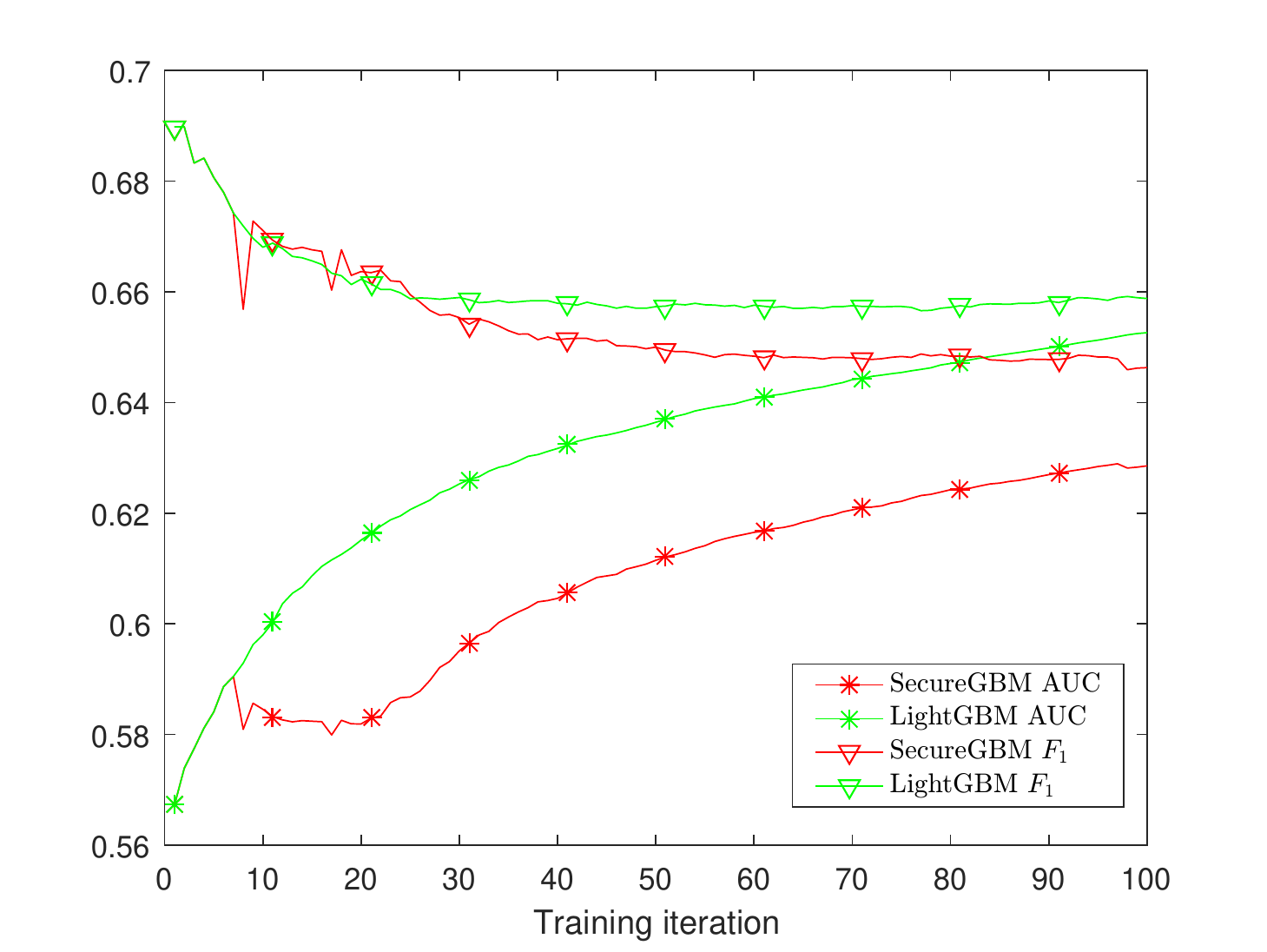}}
    \subfloat[$t=3$ on Phishing]{\includegraphics[width=0.3\textwidth]{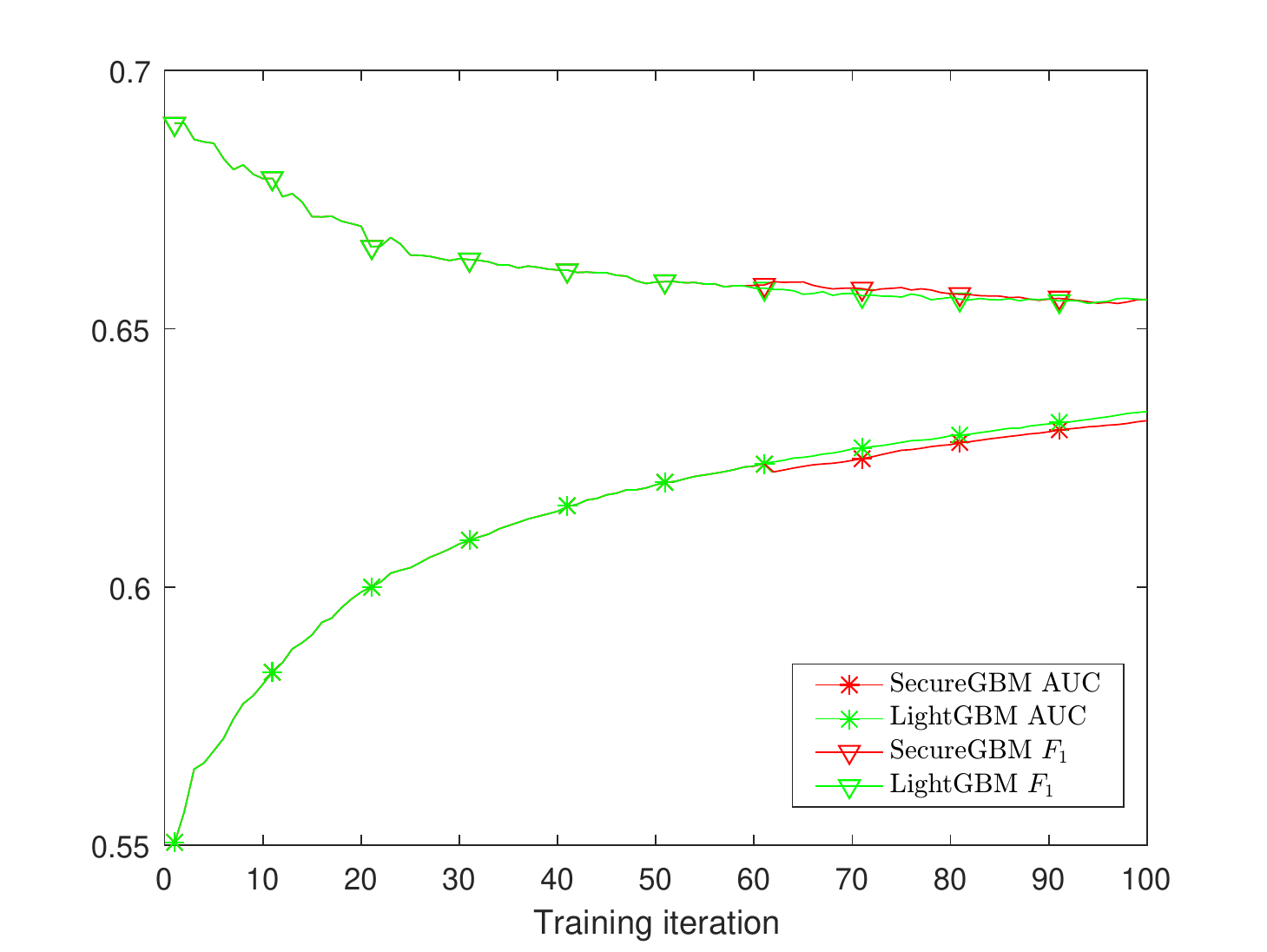}}
    \caption{The Comparison of Training AUC and F1-Score per Iteration \TheName\ vs. LightGBM-($\mathbb{A}$,$\mathbb{B}$): : Labels in sparse datasets (personal bankruptcy status) are imbalanced with most samples negative; in this case, the learned models are usually imbalanced with very low recall and F1-score.}
    \label{fig:runtime}
\end{figure*}

\begin{figure*}
    \centering
    \subfloat[$t=5$ on Sparse]{\includegraphics[width=0.3\textwidth]{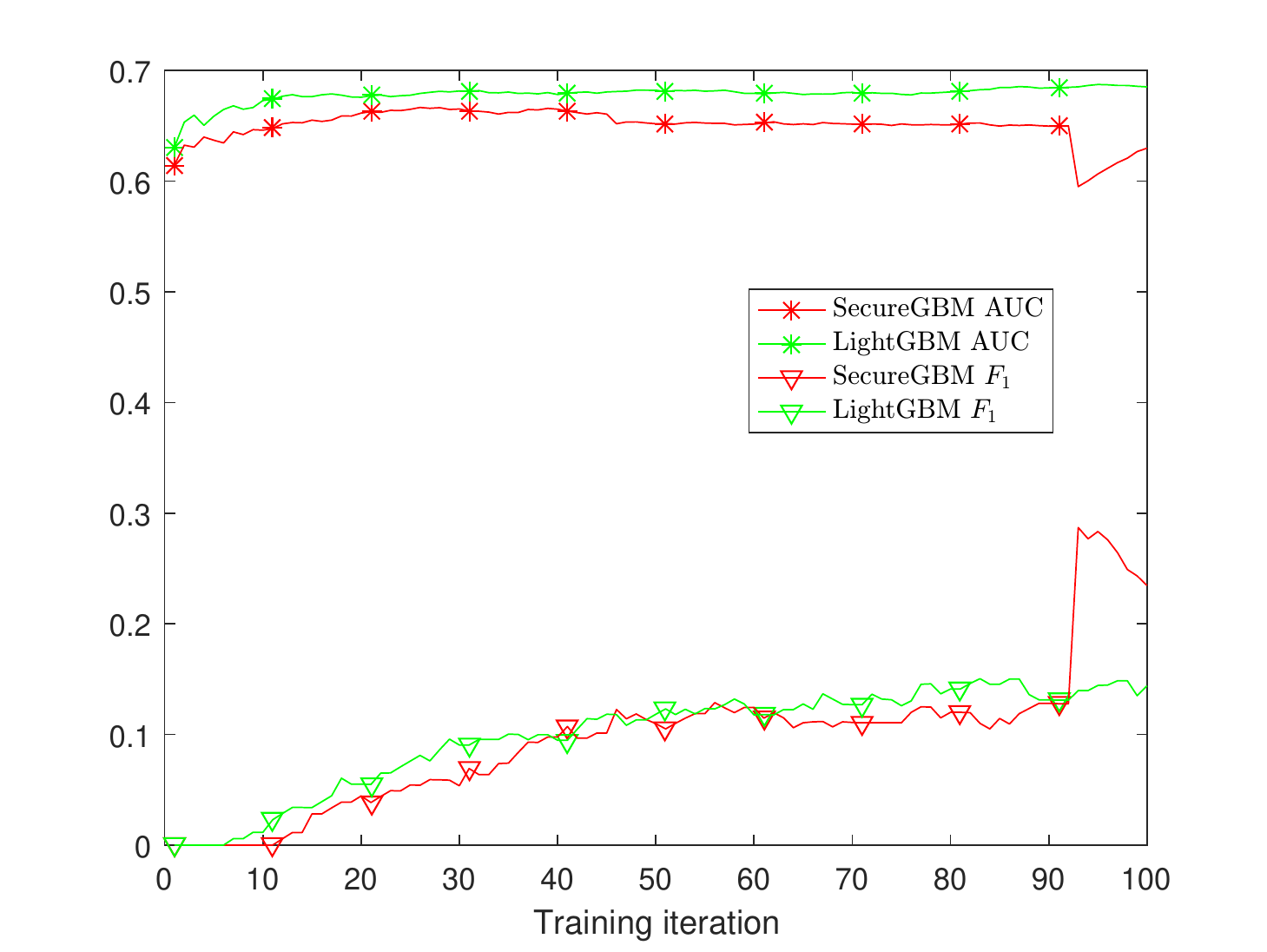}}
    \subfloat[$t=4$ on Sparse]{\includegraphics[width=0.3\textwidth]{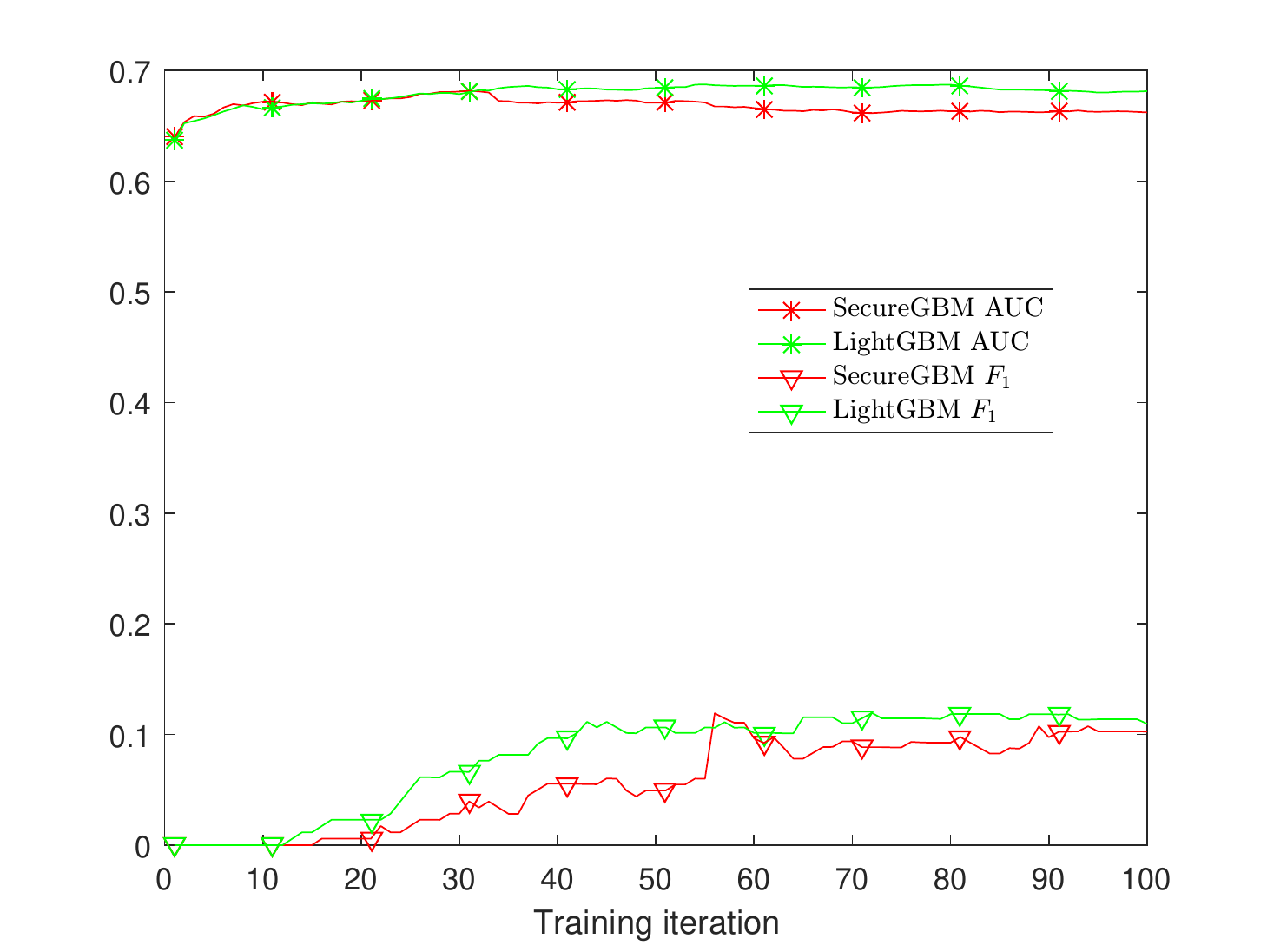}}
    \subfloat[$t=3$ on Sparse]{\includegraphics[width=0.3\textwidth]{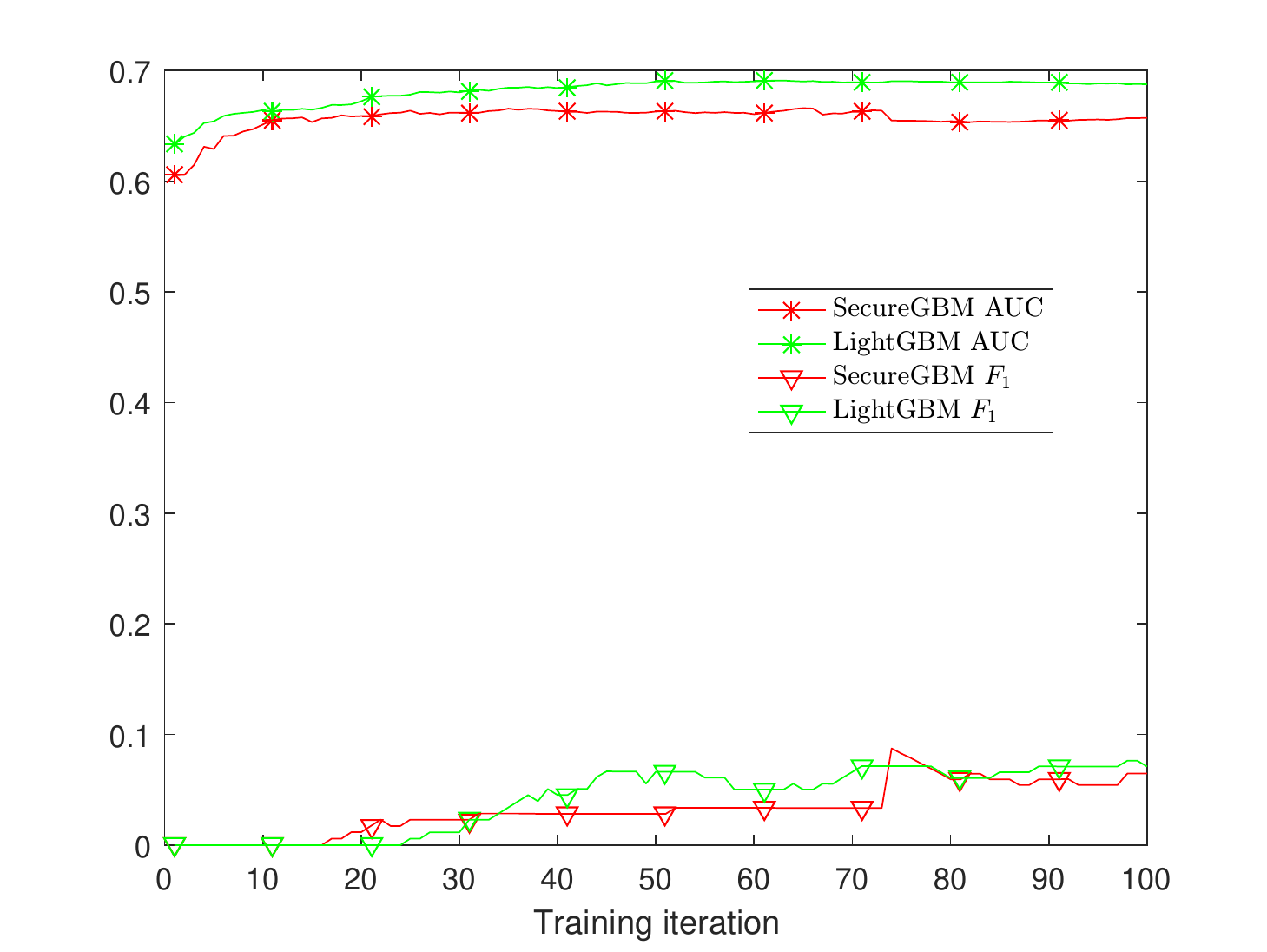}}\\
    \subfloat[$t=5$ on Adult]{\includegraphics[width=0.3\textwidth]{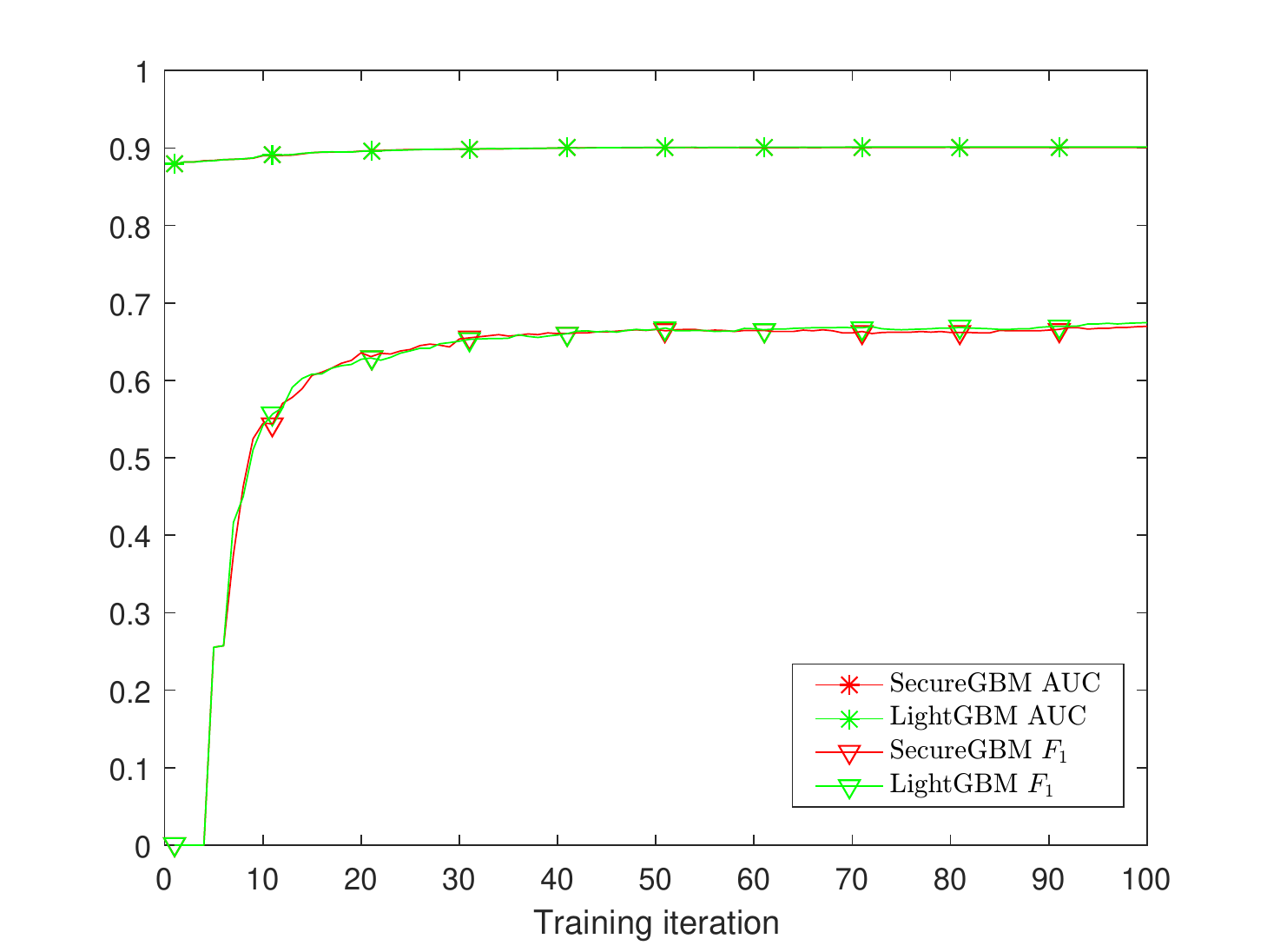}}
    \subfloat[$t=4$ on Adult]{\includegraphics[width=0.3\textwidth]{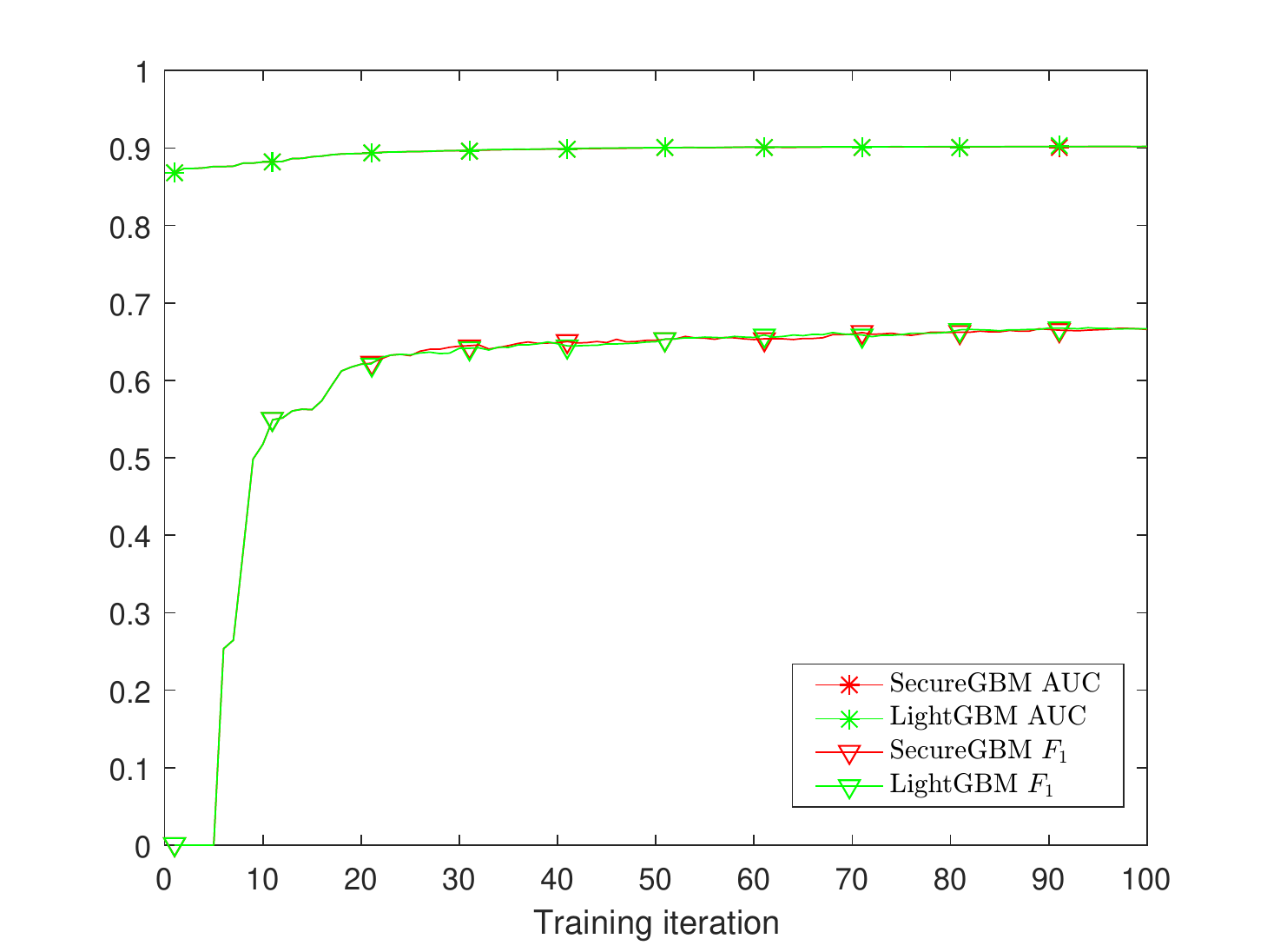}}
    \subfloat[$t=3$ on Adult]{\includegraphics[width=0.3\textwidth]{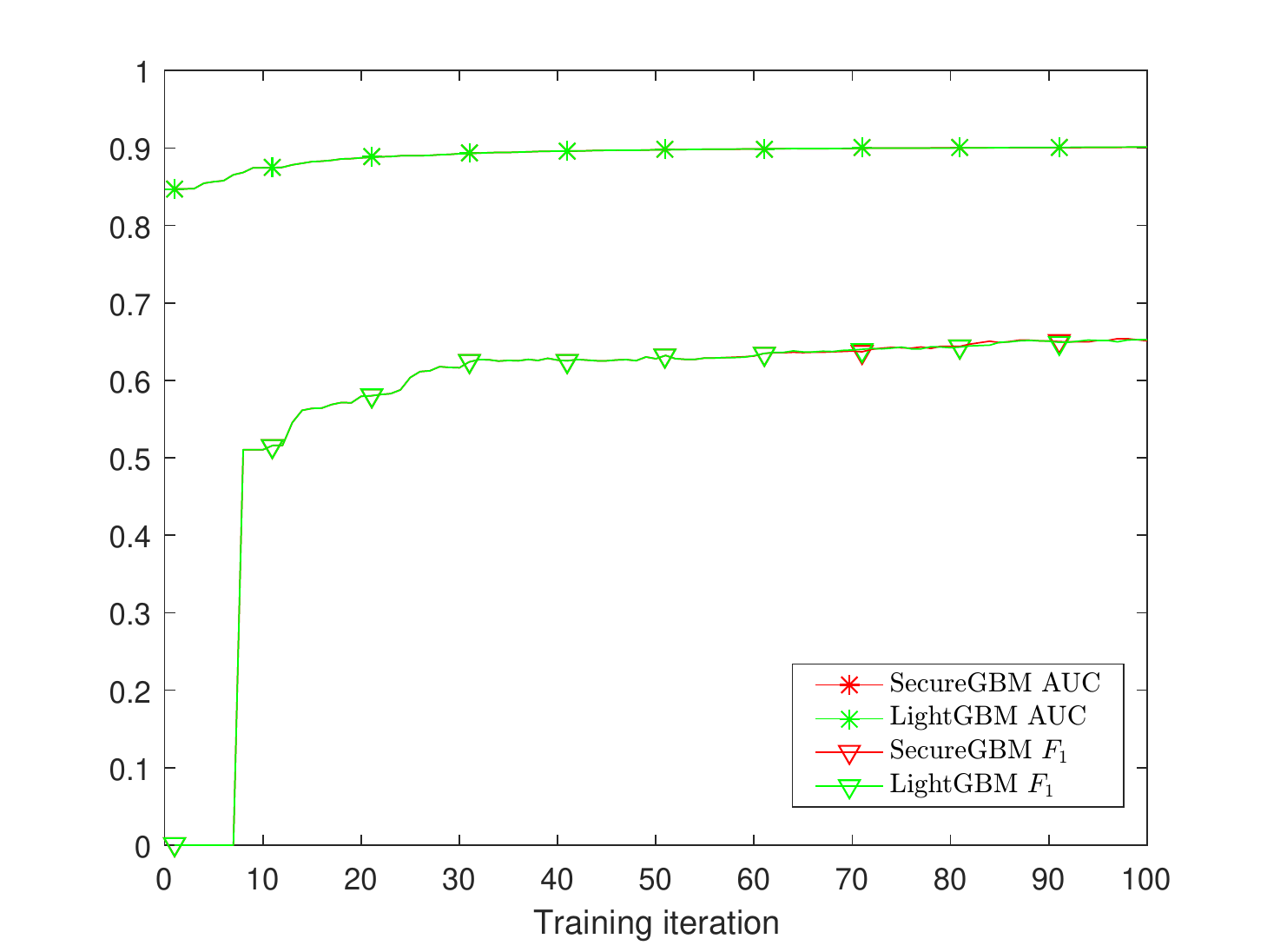}}\\
    \subfloat[$t=5$ on Phishing]{\includegraphics[width=0.3\textwidth]{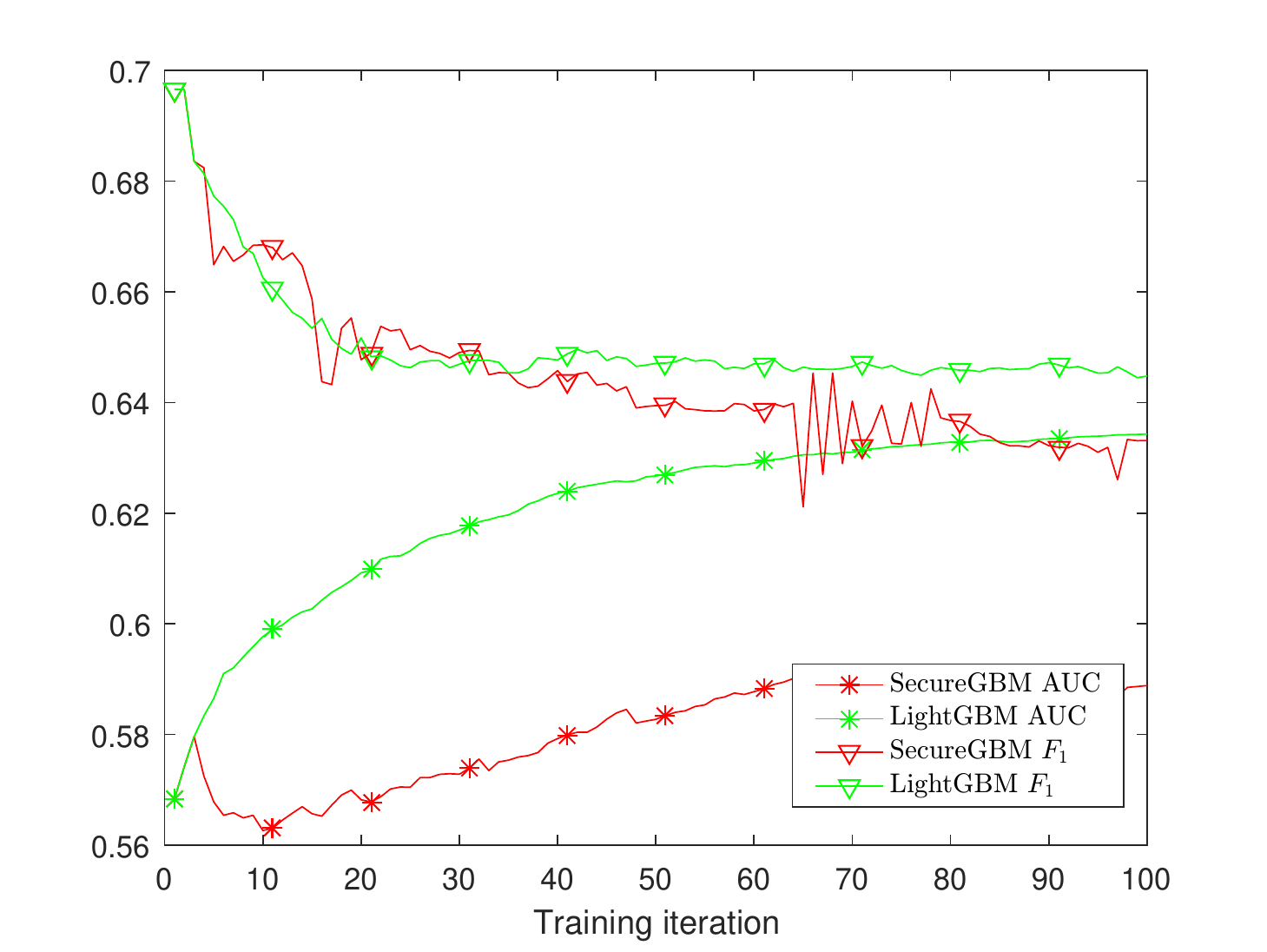}}
    \subfloat[$t=4$ on Phishing]{\includegraphics[width=0.3\textwidth]{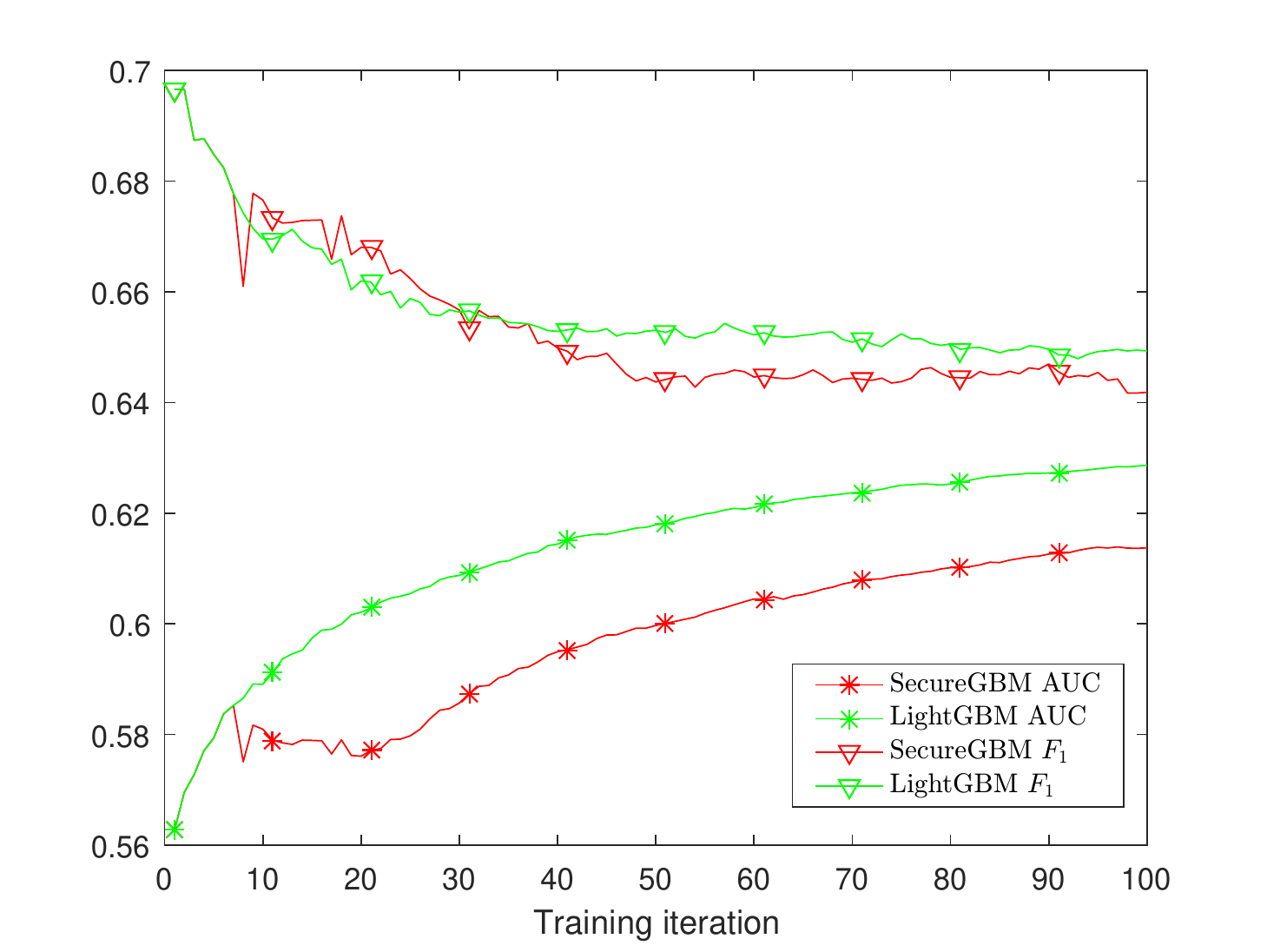}}
    \subfloat[$t=3$ on Phishing]{\includegraphics[width=0.3\textwidth]{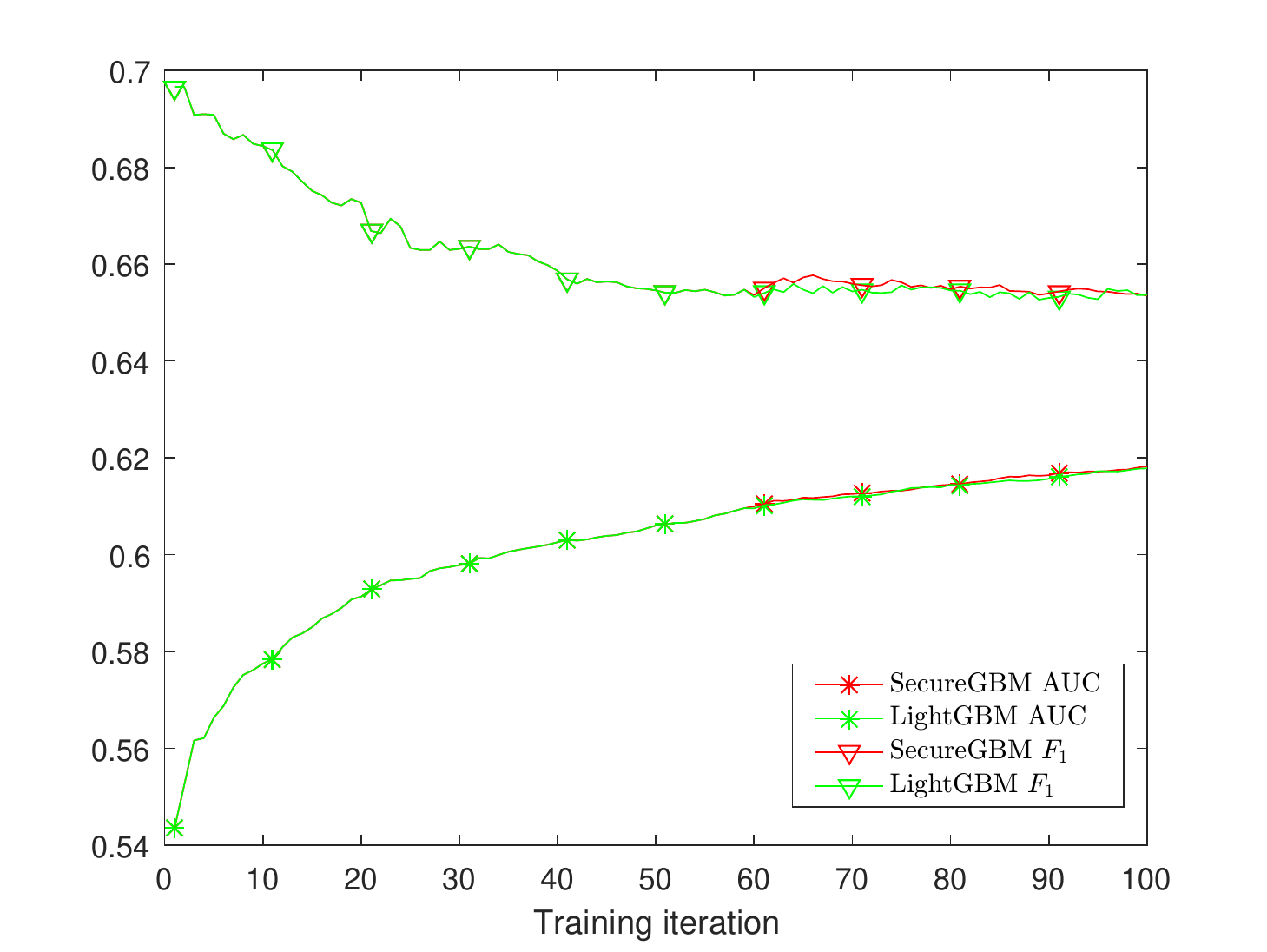}}
    \caption{The Comparison of Testing AUC and F1-Score per Iteration \TheName\ vs. LightGBM-($\mathbb{A}$,$\mathbb{B}$): : Labels in sparse datasets (personal bankruptcy status) are imbalanced with most samples negative; in this case, the learned models are usually imbalanced with very low recall and F1-score.}
    \label{fig:testing}
\end{figure*}

\subsection{Overall Performance}
To evaluate the overall accuracy of \TheName, we measured the Area Under Curve (AUC) of the \TheName\ prediction results and compared to the baseline algorithms. Table~\ref{tab:overall} presents the overall comparison on AUC achieved by these classifiers under the same settings. \TheName, LightGBM, and LIBSVM all have been trained using 200 iterations, where we measure the AUC on training and testing datasets.

The result in Table~\ref{tab:overall} shows that, compared to LightGBM-($\mathbb{A}$,$\mathbb{B}$), \TheName\ achieved similar training and testing AUC based on the same settings, while significant outperforming LightGBM-$\mathbb{A}$ that used the single dataset of $\mathbb{A}$. Furthermore, under both settings, LightGBM performed better than XGBoost in terms of testing AUC. Similar observations can be obtained through the comparisons between LIBSVM and LightGBM. In short, it is reasonable to conclude that multi-party gradient boosting over two distributed datasets can significantly improve the performance and outperforms the one that uses datasets from the party $\mathbb{A}$ only. 

Furthermore, the comparison to LightGBM-$\mathbb{B}$* shows that, except the experiments based on Sparse datasets, \TheName\ significantly outperforms the one that aggregates the features from $\mathbb{B}$ and labels from $\mathbb{A}$. For the Sparse datasets, one can easily observe that LightGBM-$\mathbb{A}$ failed to train the model, when using the datasets on $\mathbb{A}$ only, as the features in $\mathbb{A}$ are too sparse to learn. The comparison between LightGBM-($\mathbb{A}$,$\mathbb{B}$) and  LightGBM-$\mathbb{B}$*  further demonstrates that the incorporation of the features in $\mathbb{A}$ can not improve the performance of LightGBM learning. Due to the same reason, \TheName\ performed slightly worse than  LightGBM-$\mathbb{B}$* with marginal testing AUC degradation. 

We conclude \TheName\ boosts the testing accuracy of learners from the party $\mathbb{A}$ perspectives, as (1) \TheName\ consistently outperforms LightGBM-$\mathbb{A}$,  XGBoost-$\mathbb{A}$ and other learners that uses datasets on $\mathbb{A}$ only; (2) The algorithms that aggregates datasets from the both sides, such as LightGBM-($\mathbb{A}$, $\mathbb{B}$) or  LightGBM-$\mathbb{B}$*,  only perform marginally better than \TheName, while these algorithms scarifying the data privacy of the two parties.

\subsection{Case Studies}
To further understand the performance of \TheName, we traced back the the models obtained after each iteration and analysis their accuracy from both accuracy and efficiency perspectives.

\subsubsection{Trends of Accuracy Improved per Iteration} Figure~\ref{fig:runtime} presents the the comparison of training AUC and F1-score per iteration, between \TheName\ versus vanilla LightGBM. More specific, we evaluated the performance when $t=3,\ 4,\ $ and $5$ (as the LightGBM and \TheName\ use leaf-wise growth strategy, $t$ is equivalent to the depth of each decision tree learned), where we clearly observed the error convergence of the model.  

It has been observed that, in most cases, the training F1-score could be gradually improved with the increasing number of iterations. For sparse and Adult datasets, the overall trends of AUC and F1-score for LightGBM and \TheName\ were almost same under all settings --- even though, for Sparse dataset, \TheName\ only used 1\% training data as the mini-batch for the model update per iteration (while LightGBM used the whole). Furthermore, even though \TheName\ did not perform as good as LightGBM for the Phishing dataset when $t=5$ and $4$, it still achieved decent performance like LightGBM under the appropriately setting $t=3$.  Such observations are quite encouraging that the use of mini-batch seems to not hurt the learning progress of \TheName\ for these datasets, with appropriate settings. The lines of \TheName\ are with more jitters, due to the use of stochastic approximation for statistical acceleration. Similar observations have been obtained in the comparison of testing AUC and F1-score per iteration which has been shown in Figure~\ref{fig:testing}.

\begin{table}[]
    \centering
        \caption{Time Consumption per Iteration (seconds) on a Synthesized Dataset over Varying Number of Training Samples}
        \normalsize{
    \begin{tabular}{r|ccccc} \hline
      \# Samples &  1,000  & 2,000  &  4,000  & 8,000  & 16,000  \\ \hline
     \TheName &  10.20  &  10.50  &  11.4  & 11.75  &  14.30  \\ 
     LightGBM &  0.16   &  0.32   &  0.70  & 1.41   &  2.45  \\ 
     XGBoost  &  0.51   &  0.73   &  1.09  & 2.20   &  4.30  \\ \hline
    \end{tabular}}
    \label{tab:time}
    \vspace{-3mm}
\end{table}

\subsubsection{Time Consumption over Scale of Problem} To test the time consumption of \TheName\ over varying scale of the problem, we synthesize a dataset based on Sparse with increasing number of samples. The experiment results showed that the \emph{time consumption per iteration} for the training procedure of \TheName\ is significantly longer than LightGBM and XGBoost. 

We estimate the slowdown ratio of \TheName\ as the ration between the time consumption per iteration for \TheName\ versus vanilla LightGBM.  The range of slowdown ratio is around 3x$\sim$ 64x in this experiment. Furthermore, with the number of samples increases, the slowdown ratio of \TheName\ would decrease significantly. For example the ratio is around 63.75x when comparing \TheName\ to LightGBM with 1,000 training samples, while it is only 5.8 when compared to LightGBM with 16,000 training samples. It is not difficult to conclude that \TheName\ is quite time efficient due to its statistical acceleration strategies used, \TheName\ would become more and more efficient when the scale of training set increases. The experiments are carried out using two workstations based on 8-core Xeon CPUs and 16GB memory. The two machines are interconnected with a 100 MBit cable with 1.55ms latency.
\section{Discussion and Conclusion}
In this work, we present \TheName\ a secure multi-party (re-)design of LightGBM~\cite{ke2017lightgbm}, where we assume the view (i.e., set of features) of the same group of samples has been split into two parts and owned by two parties separately. To collaboratively train a model while preserving the privacy of the two parties, a group of partial homomorphic encryption (PHE) computation models and multi-party computation protocols have been used to cover the key operators of distributed LightGBM learning and inference over two parties. As the use of PHE and multi-party computation models cause hudge computational and communication overhead, certain statistical acceleration strategies have been proposed to lower the cost of communication while securing the statistical accuracy of learned model through stochastic approximation. With such statistical acceleration strategies, \TheName\ would become more and more efficient, with decreasing slowdown ratio, when the scale of training datasets increases. 

The experiments based on several large real-world datasets show that \TheName\ can achieve decent testing accuracy (i.e., AUC and F1-score) as good as vanilla LightGBM (based on the aggregated datasets from the two parties), using a time consumption tolerable training procedure (5x$\sim$ 64x slowdown), without compromising the data privacy. Furthermore, the ablation study that compares \TheName\ to the learners, which uses the single dataset from one party, showed that such collaboration between two parties can improve the accuracy.

{
\bibliographystyle{IEEEtran}
\bibliography{biblio,main}
}

\end{document}